\documentclass{article}

\usepackage{microtype}
\usepackage{graphicx}
\usepackage{subcaption}
\usepackage{booktabs} 
\usepackage[table]{xcolor} 

\usepackage{makecell}
\usepackage{enumitem}
\usepackage{multirow}

\usepackage{hyperref}
\usepackage{makecell}
\usepackage{xcolor}
\definecolor{myblue}{rgb}{0.933, 0.933, 0.996}

\usepackage[dvipsnames]{xcolor} 
\usepackage[most]{tcolorbox}

\usepackage[accepted]{icml2026}

\usepackage{times}
\usepackage{latexsym}
\usepackage[T1]{fontenc}
\usepackage[utf8]{inputenc}
\usepackage{microtype}
\usepackage{inconsolata}

\usepackage{amsmath}
\usepackage{amssymb}
\usepackage{graphicx}
\usepackage{booktabs} 
\usepackage{multirow}
\usepackage{adjustbox}
\usepackage{makecell}
\usepackage{rotating}  
\usepackage{float}    
\usepackage{caption}
\usepackage{enumitem}  
\usepackage{CJKutf8}  

\usepackage{tcolorbox}

\definecolor{lightblue}{RGB}{100, 100, 255}
\definecolor{lightred}{RGB}{255, 100, 100}

\tcbset{
    promptbox/.style={
        colframe=black,
        colback=white,
        boxrule=1pt,
        title={Prompt Template}, 
        coltitle=white,
        fonttitle=\bfseries,
        left=6pt, right=6pt, top=6pt, bottom=6pt,
        colbacktitle=RedViolet,
        arc=2pt 
    }
}

\usepackage{inconsolata}

\usepackage{graphicx}
\usepackage{fontawesome5}
\usepackage{tikz}
\usepackage{subcaption}
\usepackage{arydshln}
\usepackage{capt-of}

\usepackage{amsmath}
\usepackage{amssymb}
\usepackage{mathtools}
\usepackage{amsthm}

\usepackage[capitalize,noabbrev]{cleveref}

\theoremstyle{plain}

\theoremstyle{definition}

\theoremstyle{remark}

\usepackage[textsize=tiny]{todonotes}

\icmltitlerunning{ICML 2026 - AI for Law}

\begin{document}

\twocolumn[
  \icmltitle{{TransLaw: A Large-Scale Dataset and Multi-Agent Benchmark \\Simulating Professional Translation of Hong Kong Case Law}}

  \icmlsetsymbol{equal}{*}

  \begin{icmlauthorlist}
    \icmlauthor{Xi Xuan}{cityu}
    \icmlauthor{Chunyu Kit}{cityu}
  \end{icmlauthorlist}

  \icmlaffiliation{cityu}{City University of Hong Kong, Hong Kong SAR, China}

  \icmlcorrespondingauthor{Xi Xuan}{xixuan3@cityu.edu.hk}
  \icmlcorrespondingauthor{Chunyu Kit}{ctckit@cityu.edu.hk}

  \icmlkeywords{Machine Learning, ICML}

  \vskip 0.3in
]

\printAffiliationsAndNotice{}  

\begin{abstract}

Translating Hong Kong Court Judgments from English to Traditional Chinese is mandated by Articles 8–9 of the Basic Law, yet remains constrained by a shortage of parallel resources and rigorous demands on legal terminology, citation format, and judicial style. We introduce HKCFA Judgment 97-22, the first large-scale sentence-aligned parallel corpus for HK case law, comprising 344 professionally translated judgments (11,099 sentence pairs; 2.1M tokens) spanning 1997–2022. Building on this resource, we propose TransLaw, a multi-agent framework that decomposes translation into word-level expression, sentence-level translation, and multidimensional review, integrating a specialized Hong Kong legal glossary database, Retrieval-Augmented Generation, and iterative feedback, with four-dimensional expert review covering semantic alignment, terminology, citation, and style. Benchmarking 13 open-source and commercial LLMs, we demonstrate that TransLaw significantly outperforms single-agent baselines across all evaluated models, with convergence within 3 iterations. Human evaluation by 10 certified legal translators using our proposed Legal ACS metric confirms gains in legal-semantic accuracy, while showing that TransLaw still trails human experts in stylistic naturalness. The dataset and benchmark code are available at https://github.com/xuanxixi/TransLaw.
\end{abstract}
\vspace{-0.85cm}
\section{Introduction}

\begin{quote}
    \itshape 
    Legal language is a specialized type of language that, while derived from ordinary language, is often much more formulaic and complex.
    \par 
    \hfill 
    \normalfont --- ~\cite{tiersma1999legal}
\end{quote}

\begin{figure*}[t]
    \centering
    \includegraphics[width=0.99\textwidth]{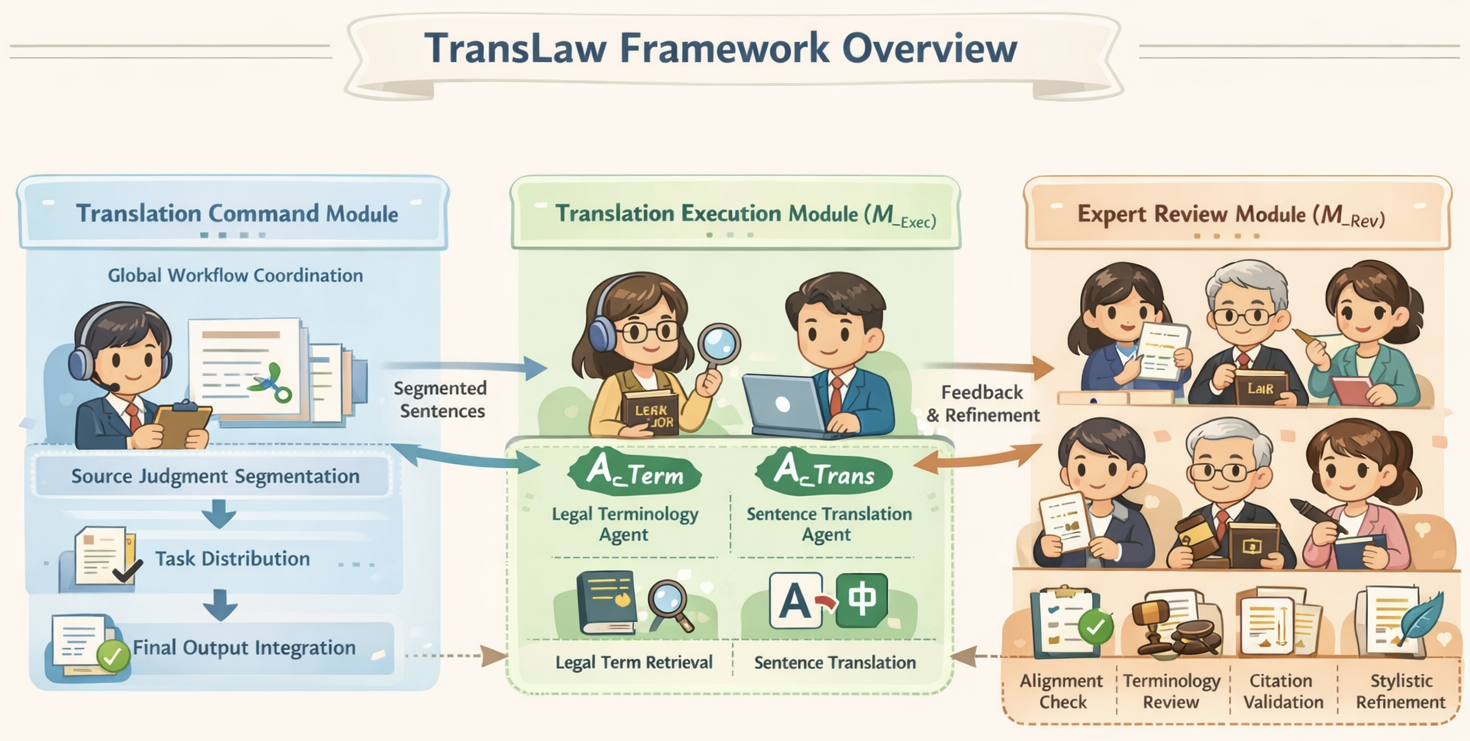}
    \caption{\small The overall architecture of TransLaw. The framework consists of three collaborative modules: (1) the Translation Command Module ($\mathcal{M}_{\text{Com}}$), where $\mathcal{A}_{\text{Com}}$ coordinates the global workflow; (2) the Translation Execution Module ($\mathcal{M}_{\text{Exec}}$), comprising $\mathcal{A}_{\text{Term}}$ and $\mathcal{A}_{\text{Trans}}$ for legal terminology parsing and core translation; and (3) the Expert Review Module ($\mathcal{M}_{\text{Rev}}$), which integrates $\mathcal{A}_{\text{Align}}$, $\mathcal{A}_{\text{TermR}}$, $\mathcal{A}_{\text{Cita}}$, and $\mathcal{A}_{\text{StyleP}}$ for multi-dimensional quality verification.}
    \vspace{-0.42cm}
    \label{fig:2}
\end{figure*}Judgment translation is the focal point of this research, which aims not only to bridge the critical research gap in Machine Translation (MT) for Hong Kong (HK) Case Law but also, more importantly, to meet a particular practical social need with tremendous potential for the Hong Kong legal system. Following the return of Hong Kong to China on 1 July 1997 after 155 years of British rule, the Hong Kong Special Administrative Region (HKSAR) took pride, upon its establishment, in having all its written statutes translated into Chinese, a mammoth task once considered impossible by many. This was required by the bilingual legislation as was mandated by Articles 8 and 9 of the Basic Law,\footnote{{\scriptsize \url{https://www.basiclaw.gov.hk/en/basiclaw/chapter1.html}}} the former stipulating the retention of the English common law system, and the latter establishing legal bilingualism with Chinese and English as official languages, which are interpreted as being equally authentic\footnote{{\scriptsize \url{https://www.elegislation.gov.hk/hk/cap1!en/s10B}}} or of equal status.\footnote{{\scriptsize \url{https://www.doj.gov.hk/en/about/orgchart_ldd_drafting_chi_eng.html}}} Against this constitutional backdrop, the translation of Hong Kong judgments constitutes a pivotal component in sustaining the territory's bilingual legal framework \cite{Cheng2016}. Despite persistent challenges in reconciling linguistic transformation with inherited legal infrastructure \cite{Chen2002}, the system traces back to the 1987 Bilingual Laws Project, which systematized statute translation \cite{Jones1987} and institutionalized parallel legislative drafting \cite{Mushkat1997}. While English remained the predominant courtroom language post-handover \cite{daniels2011legacy}, progressive legal localization \cite{tam2012legal} has rendered judgment translation an essential mechanism ensuring jurisprudential precision \cite{PrietoRamos2014} and facilitating cross-jurisdictional legal communication.

Confronted with the voluminous common law documentation within HK's judicial system \cite{Hau2019}, the establishment of efficient, accurate, and large-scale translation processes for laws assumes critical significance in HK \cite{sin2026solving}. However, launching another project like the 1987 one for translating case law texts totally by experts' manual work is unrealistic. Even if such an undertaking might be conceived as possible for the existing case law, this manual approach could hardly stand a chance of dealing with the innumerable upcoming cases in the future. Thus, shifting from exclusively manual workflows to a hybrid paradigm that integrates mainstream Large Language Models (LLMs) with professional human translation offers a promising direction. 
Significant stylistic differences exist in judgments across the hierarchy of Hong Kong courts. Judgments from the Court of Appeal, the Court of First Instance, the District Court, and other lower courts tend to be stylistically flexible. In contrast, the Court of Final Appeal (HKCFA) are characterized by strict constraints on format, structure, syntax, and terminology \cite{chan2012bridging, ng2020common}. Thus, this study targets HKCFA judgment translation, representing a more challenging legal language domain.

Previous research has successfully explored machine translation in the legal domain. However, these studies primarily focus on documents from various jurisdictions and language pairs ~\cite{bajcic2024applying, niklaus2025swiltra, singh2025evaluation,li2025legalagentbench}, which differ significantly from HKCFA judgments. Recent research provides a systematic legal MT overview of the shift from traditional Neural Machine Translation (NMT) to Transformer-based Language Models (TLMs)~\cite{greco2024bringing}.

Another study compared the differences between
machine translation and human translation of English legal texts into  Arabic ~\cite{moneus2024artificial}. The authors concluded that machine translation is not suitable for translating English legal texts into Arabic as it fails to adhere to strict formatting constraints or comprehend the socio-cultural background and complex legal terminology necessary to guarantee 'legal effect,' resulting in poor translation quality.

LLMs, based on the Transformer architecture with billions to hundreds of billions of parameters, have demonstrated their effectiveness in machine translation across
various language pairs~\cite{treviso2024xtower,elshin2024general, chen2025llm,feng2025tear}. Consequently, recent research has begun exploring the potential of LLMs for legal translation focusing on both specialized model tuning and the creation of robust evaluation benchmarks.  For instance, recent research examined the effectiveness  of  evaluation of frontier LLMs and finetuned open small language models (SLMs) on Swiss legal translations in both zero-shot and fine-tuning settings in translating Swiss Legal document and found the results to be unsatisfactory~\cite{niklaus2025swiltra}. This research highlighted several issues with the text generated by LLM, including inappropriate legal word choices, hallucinating non-existent terms, and failing to grasp complex legal terminology. Furthermore, the model struggles to comprehend the socio-cultural background and cultural nuances embedded in legal systems, and fails to adhere to strict formatting constraints and formalized structural conventions required for official legal proceedings.

While previous work has advanced the field by enhancing models and establishing evaluation benchmarks, it has primarily focused on improving the intrinsic capabilities of single LLMs. Consequently, the potential of current mainstream general and legal-specific LLMs for translating Hong Kong legal judgments remains unclear and unexplored. To address the shortage of bilingual Hong Kong legal data and advance legal translation, we present four main contributions:

\begin{enumerate}[leftmargin=*, labelsep=0.5em, itemsep=0pt, parsep=0pt, topsep=0pt]
    \item We are the first to examine the capabilities of LLMs in English-to-Chinese HK Case Law translation.
    
    \item We construct and release the HKCFA Judgment 97-22 dataset, the first high-quality dataset comprising sentence-level parallel pairs of judicial judgments.

    \item We propose a novel multi-agent judgment translation system that integrates a Hong Kong legal glossary, RAG, and iterative feedback. By decomposing translation into word-level expression, sentence-level translating, and multidimensional review, our approach significantly outperforms single-agent systems.
    
    \item We propose a new human-evaluation metric, the Legal ACS metric, for HK judgment translation, and engage professional translators with expertise in HK legal translation to evaluate the results.
\end{enumerate}

\vspace{-10pt}

\section{Related Work}

\paragraph{Multi-Agent Systems (MAS)} These systems consist of multiple autonomous yet collaborating agents. Characterized by core features such as autonomy, interactivity, cooperativity, and distribution, they can tackle challenges beyond the reach of single agents \cite{guo2024multiagents}. The rapid evolution of LLMs has invigorated MAS, demonstrating its significant potential in various fields such as software development \cite{hong2023metagpt}, multi-robot collaboration \cite{mandi2024roco}, scientific experiments \cite{du2023multiagentdebate}, and scientific debates \cite{xiong2023interconsistency}. Moreover, LLM-based MAS play a crucial role in world simulation for social sciences, psychology, economics, hospital and scientific discovery, (re)enacting various roles and perspectives through agents’ role-playing \cite{tao2025magis,yu2025fincon,he2025llm,benita2025phoenix,yao2025survey,ghafarollahi2025sciagents, liu2026dynamic}.
\vspace{-10pt}

\paragraph{MAS in Machine Translation}
HK judgment translation imposes rigid terminological and structural
demands, and multi-agent collaboration is well suited to handling such
complexity. Yet MAS-based translation has received little attention in
legal NLP. Inspired by recent advances in multi-agent collaboration for
translation, Liang et al.~\cite{liang-etal-2024-encouraging} proposed the
Multi-Agent Debate (MAD) framework, which improves performance on tasks
like machine translation by encouraging diverse reasoning through
structured argumentation. Wu et al.~\cite{wu2024beyond} proposed
TransAgents, which focuses on literary translation by incorporating
role-based agents such as translator, editor, and proofreader.
Lv et al.~\cite{lv2025multi} leveraged MAS for Classical Chinese
translation, adopting a modular architecture with dedicated agents for
keyword interpretation and grammatical validation to ensure cultural
fidelity and semantic accuracy. These studies underscore the
effectiveness of MAS in translation, yet they are confined either to
translation evaluation under the LLM-as-a-judge paradigm or to
literary and classical adaptation centered on figurative language,
cultural nuance, and stylistic fidelity. We are the first to propose a
MAS-based method for HK judgment translation, targeting legal
terminology accuracy, culturally embedded nuances, and strict linguistic
structures, which is detailed in the
next section.

\vspace{-10pt}
\section{Methodology}

\subsection{HKCFA Judgment 97-22 Dataset}

A high-quality evaluation dataset is a prerequisite for advancing research on translating HK Case Law by LLMs~\cite{fei2025internlm}. Currently, there is no publicly available dataset specifically designed for the bilingual translation of HK Case Law. To address this gap, we constructed a novel dataset named HKCFA Judgment 97-22, derived from Chinese-English bilingual \textbf{H}ong \textbf{K}ong \textbf{C}ourt of \textbf{F}inal \textbf{A}ppeal \textbf{Judgment}s from 19\textbf{97} to 20\textbf{22}. The dataset includes 344 high-quality judgments produced by professional human translation, with detailed statistics presented in Table~\ref{tab:hkcfa_stats_detailed}.

\begin{table}[!h]
    \centering
    \caption{\footnotesize The statistics of HKCFA Judgment 97-22 Dataset.}
    
    \setlength{\heavyrulewidth}{1.4pt}
    \setlength{\lightrulewidth}{0.5pt}

    \resizebox{\columnwidth}{!}{%
        \begin{tabular}{lccc} 
            \Xhline{1.4pt}
            \textbf{Word-Count} & \textbf{Sentences} & \textbf{En-Token} & \textbf{Zh-HK-Token} \\ 
            \midrule
            5--15    & 2,982  & 28,541  & 56,573   \\
            16--35   & 1,998  & 65,842  & 108,872  \\
            36--54   & 1,631  & 92,714  & 152,428  \\
            55--100  & 2,553  & 239,978 & 390,492  \\
            101--200 & 1,599  & 271,173 & 433,178  \\
            201--702 & 336    & 113,405 & 172,739  \\
            \midrule
            \textbf{Total} & \textbf{11,099} & \textbf{811,653} & \textbf{1,314,282} \\ 
            \Xhline{1.4pt}
        \end{tabular}%
    }
    
    \label{tab:hkcfa_stats_detailed}
\end{table}

To calculate these statistics, we use the tiktoken ~\cite{jain2025tiktoken} with the corresponding Byte-Pair Encoding (BPE) token encoding, which is
also used by GPT-4o mini ~\cite{menick2024gpt}. While existing parallel legal corpora often use automated methods for sentence alignment~\cite{koehn2005europarl, ziemski2016united}, our dataset relies on the structure provided by official government bodies, such as law paragraphs embedded in the HTML, resulting in high-quality alignment. To the best of our knowledge, this constitutes the first Hong Kong Case Law dataset comprising sentence-level parallel data pairs of Chinese-English judgments derived from professional human translation. 
\vspace{-10pt}
\subsection{TransLaw Framework Overview}
Translating English legal judgments into Traditional Chinese effectively is complex. It requires not only accurately grasping legal terminology and preserving the socio-cultural depth embedded in legal systems, but also strictly adhering to the formalized structural conventions required for official legal proceedings. To address these challenges, this study constructs a multi-agent translation system, TransLaw, for judgment translation.

The overall translation task is formulated as processing source judgment $J$, initially segmented into a sequence of sentences based on semantic structure, defined as $J = \{s_i \mid i = 1, 2, \dots, N\}$. Each sentence $s_i$ serves as an independent unit for bilingual processing by the translation system. TransLaw functions through three modules operated by a set of role agents:

\begin{enumerate}[leftmargin=*, nosep]
    \item \underline{\textit{Translation Command Module}} {\small ($\mathcal{M}_{\text{Com}}$)} houses Translation Command Agent ($\mathcal{A}_{\text{Com}}$), acting as central task coordinator responsible for sentence segmentation, information dispatch, and result integration.
    \item \underline{\textit{Translation Execution Module}} {\small ($\mathcal{M}_{\text{Exec}}$)} comprises Legal Terminology Agent ($\mathcal{A}_{\text{Term}}$) and Sentence Translation Agent ($\mathcal{A}_{\text{Trans}}$). The former focuses on retrieving and optimizing interpretations for legal terms, while the latter generates initial translations based on these interpretations and surrounding context.
    \item \underline{\textit{Expert Review Module}} {\small ($\mathcal{M}_{\text{Rev}}$)} integrates four verification agents to ensure final quality. First, Semantic Alignment Agent ($\mathcal{A}_{\text{Align}}$) performs fine-grained cross-checks between source and target texts to guarantee strict logical and factual consistency. Second, Legal Term Review Agent ($\mathcal{A}_{\text{TermR}}$) verifies legal terminology against the shared access authoritative HK legal glossary. Third, Legal Citation Agent ($\mathcal{A}_{\text{Cita}}$) validates that all case references and legislative provisions adhere to rigid HKCFA formatting standards. Fourth, Stylistic Fidelity Polishing Agent ($\mathcal{A}_{\text{StyleP}}$) refines linguistic output, ensuring translation exhibits appropriate judicial tone and syntactic fluency.
\end{enumerate}

In summary, the TransLaw Framework formally defines the collective system of agents $\mathcal{C}$ as:
\vspace{-10pt}
\begin{equation}
\resizebox{0.95\columnwidth}{!}{%
$ \mathcal{C} = \left\{
\underbrace{\mathcal{A}_{\text{Com}}}_{\text{Command}}, 
\underbrace{\mathcal{A}_{\text{Term}}, \mathcal{A}_{\text{Trans}}}_{\text{Execution}}, 
\underbrace{\mathcal{A}_{\text{Align}}, 
\mathcal{A}_{\text{TermR}},
\mathcal{A}_{\text{Cita}}, \mathcal{A}_{\text{StyleP}}}_{\text{Review}}
\right\}. $
}
\end{equation}

Multiple agents collaborate through unified collaborative and feedback mechanisms, ensuring coherence and high quality of final translation output. Overall framework is depicted in Fig. \ref{fig:2}.

\begin{table*}[htbp]
\centering
\caption{\small The performance (\%) of various LLM models serving as agents in the TransLaw multi-agent framework compared to a Single Translator Agent. Best results are in \textbf{bold}. Confidence intervals are in parentheses.For the DeepSeek family, \textit{MoE-16B} refers to DeepSeekMoE-16B-Base and \textit{R1-Distill-32B} refers to DeepSeek-R1-Distill-Qwen-32B (see Table~\ref{tab:llm} for version details).}
\renewcommand{\arraystretch}{1.15} 
\setlength{\tabcolsep}{6pt}       

\resizebox{\textwidth}{!}{%
\footnotesize 
\begin{tabular}{ll|ccc|ccc|c}
\Xhline{1.4pt}
\multirow{2}{*}{\textbf{Series}} & \multirow{2}{*}{\textbf{Model}} &
\multicolumn{3}{c}{\textbf{TransLaw}} & \multicolumn{3}{c}{\textbf{Single Translator Agent}} & \multirow{2}{*}{\textbf{Rank}} \\
\cmidrule(lr){3-5} \cmidrule(lr){6-8}
& & xCOMET-XL$\uparrow$ & wmt22-unite-da$\uparrow$ & \textbf{Avg.}$\uparrow$ & xCOMET-XL$\uparrow$ & wmt22-unite-da$\uparrow$ & \textbf{Avg.}$\uparrow$ & \\
\Xhline{1.4pt}

\multirow{3}{*}{OpenAI} 
 & \cellcolor{myblue}GPT-4o        & \cellcolor{myblue}\textbf{85.12} {\scriptsize($\pm$0.14)} & \cellcolor{myblue}\textbf{91.78} {\scriptsize($\pm$0.12)} & \cellcolor{myblue}\textbf{88.45} & \cellcolor{myblue}\textbf{69.42} {\scriptsize($\pm$0.15)} & \cellcolor{myblue}\textbf{75.88} {\scriptsize($\pm$0.13)} & \cellcolor{myblue}\textbf{72.65} & \cellcolor{myblue}1 \\
 & GPT-4         & 84.24 {\scriptsize($\pm$0.16)}          & 90.10 {\scriptsize($\pm$0.15)}          & 87.15          & 68.10 {\scriptsize($\pm$0.18)}          & 74.25 {\scriptsize($\pm$0.16)}          & 71.17          & 2 \\
 & ChatGPT       & 82.29 {\scriptsize($\pm$0.19)}          & 88.52 {\scriptsize($\pm$0.18)}          & 85.41          & 66.15 {\scriptsize($\pm$0.21)}          & 72.10 {\scriptsize($\pm$0.20)}          & 69.12          & 5 \\
\hdashline

\multirow{2}{*}{DeepSeek} 
 & MoE-16B            & 83.53 {\scriptsize($\pm$0.15)}          & 89.56 {\scriptsize($\pm$0.14)}          & 86.55          & 67.45 {\scriptsize($\pm$0.17)}          & 73.45 {\scriptsize($\pm$0.16)}          & 70.45          & 3 \\
 & R1-Distill-32B            & 83.25 {\scriptsize($\pm$0.17)}          & 89.33 {\scriptsize($\pm$0.15)}          & 86.29          & 67.12 {\scriptsize($\pm$0.19)}          & 73.12 {\scriptsize($\pm$0.17)}          & 70.12          & 4 \\
\hdashline

\multirow{2}{*}{Qwen} 
 & 14B-Chat      & 81.86 {\scriptsize($\pm$0.20)}          & 87.12 {\scriptsize($\pm$0.19)}          & 84.49          & 65.55 {\scriptsize($\pm$0.22)}          & 71.80 {\scriptsize($\pm$0.21)}          & 68.67          & 6 \\
 & 7B-Chat       & 80.54 {\scriptsize($\pm$0.22)}          & 87.06 {\scriptsize($\pm$0.21)}          & 83.80          & 63.90 {\scriptsize($\pm$0.25)}          & 70.15 {\scriptsize($\pm$0.23)}          & 67.02          & 7 \\
\hdashline

\multirow{2}{*}{Baichuan} 
 & 13B-Chat      & 81.33 {\scriptsize($\pm$0.23)}          & 87.51 {\scriptsize($\pm$0.22)}          & 84.42          & 64.40 {\scriptsize($\pm$0.26)}          & 70.50 {\scriptsize($\pm$0.24)}          & 67.45          & 8 \\
 & 13B-Base      & 79.60 {\scriptsize($\pm$0.25)}          & 86.51 {\scriptsize($\pm$0.24)}          & 83.06          & 62.20 {\scriptsize($\pm$0.28)}          & 69.10 {\scriptsize($\pm$0.27)}          & 65.65          & 10 \\
\hdashline

\multirow{2}{*}{ChatGLM} 
 & 3-6B          & 80.27 {\scriptsize($\pm$0.24)}          & 87.20 {\scriptsize($\pm$0.22)}          & 83.74          & 63.85 {\scriptsize($\pm$0.27)}          & 69.20 {\scriptsize($\pm$0.25)}          & 66.52          & 9 \\
 & 2-6B          & 78.11 {\scriptsize($\pm$0.28)}          & 84.78 {\scriptsize($\pm$0.26)}          & 81.45          & 61.10 {\scriptsize($\pm$0.30)}          & 67.50 {\scriptsize($\pm$0.29)}          & 64.30          & 12 \\
\hdashline

\multirow{2}{*}{ChatLaw} 
 & 33B           & 79.29 {\scriptsize($\pm$0.26)}          & 85.80 {\scriptsize($\pm$0.25)}          & 82.55          & 62.50 {\scriptsize($\pm$0.29)}          & 68.80 {\scriptsize($\pm$0.28)}          & 65.65          & 11 \\
 & 13B           & 76.26 {\scriptsize($\pm$0.30)}          & 82.59 {\scriptsize($\pm$0.29)}          & 79.43          & 58.50 {\scriptsize($\pm$0.35)}          & 64.80 {\scriptsize($\pm$0.32)}          & 61.65          & 13 \\

\Xhline{1.4pt}

\end{tabular}
}
\label{tab:1re}
\vspace{-0.3cm}
\end{table*}

\subsection{Agent Task Allocation}
This section details how agents collaborate within the three modules: the first coordinates overall workflow, the second performs core translation tasks, and the third ensures quality through multi-dimensional verification. The following subsections detail how these modules work in concert to achieve high-quality legal translation, co-working closely in a workflow highly similar to that of professional human translators.

\paragraph{Translation Command Module}
As shown in Fig. \ref{fig:2}, $\mathcal{A}_{\text{Com}}$ orchestrates TransLaw's workflow as central coordinator, performing sentence segmentation, task distribution, and iterative refinement. $\mathcal{A}_{\text{Com}}$ initiates the workflow by segmenting the source judgment $J$ into a sequence $\{s_1, \dots, s_n\}$. To ensure consistency, it maintains a memory mechanism $\mathcal{E}_i = \{\hat{s}_1, \dots, \hat{s}_{i-1}\}$, where $\hat{s}_i$ denotes the finalized translation of the preceding sentence $s_i$.The refinement process follows an iterative loop: an initial translation $\hat{s}_i^{(0)}$ is updated based on receiving review feedback $\mathcal{F}_i^{(k)}$ from the $k$-th review round. This feedback is mapped to textual increments via a mapping function $\Psi(\cdot)$ and integrated as follows:
\vspace{-8pt}
{
\setlength{\belowdisplayskip}{3pt} 
\begin{equation}
\hat{s}_i^{(k+1)} = \hat{s}_i^{(k)} \oplus \Psi(\mathcal{F}_i^{(k)})
\label{eq:iterative_update}
\end{equation}
}where $\oplus$ denotes the integration of mapped suggestions. The loop ends when $\mathcal{F}_i^{(k)} = \emptyset$ or the iteration threshold $K$ is reached, ensuring both quality and efficiency. Once the translation and review of all sentences are complete, $\mathcal{A}_{\text{Com}}$ aggregates all finalized translations to generate the final output.

\paragraph{Translation Execution Module} As shown in Fig. \ref{fig:2}, $\mathcal{M}_{\text{Exec}}$ comprises two synergistic agents: the $\mathcal{A}_{\text{Term}}$ for micro-level semantic parsing and the $\mathcal{A}_{\text{Trans
}}$ for macro-level segment reconstruction. These agents cooperate to transform English legal judgments into Chinese translations characterized by lexical accuracy, socio-cultural nuance, and standardized judicial formatting.

The $\mathcal{A}_{\text{Term}}$ focuses on disambiguating key HK legal terminology with significant juridical or semantic complexity. Utilizing Retrieval-Augmented Generation (RAG) \cite{salemi2024evaluating,li2026towards}, it extracts the set of legal terms $L_i$ from the source sentence $s_i$ and retrieves relevant expressions from the authoritative HK legal glossary database $D$ (the Combined DOJ Glossaries of Legal Terms\footnote{https://www.glossary.doj.gov.hk/}). The initial expression for a legal term $l_k \in s_i$ is obtained from database $D$ using a retrieval function $R_D$:
\vspace{-8pt}
{
\setlength{\belowdisplayskip}{3pt} 
\begin{equation}
C_k = R_D(l_k)
\end{equation}}where $C_k$ is the set of candidate legal expressions for term $l_k$. This ensures that polysemous HK legal terms are rendered correctly within the judgments.

Leveraging the $\mathcal{A}_{\text{Term}}$'s output, the $\mathcal{A}_{\text{Trans}}$ generates sentence-level translations by integrating verified legal expressions with contextual memory. Initially, it substitutes the expressions of the target vocabulary based on the optimized term set $\hat{L}_i$:
\vspace{-8pt}
{
\setlength{\belowdisplayskip}{3pt} 
\begin{equation}
s'_i = h(s_i, \hat{L}_i)
\end{equation}}where $h(s_i, \hat{L}_i)$ represents the initial transformation function injecting $\hat{L}_i$ into the syntactic structure of $s_i$. To ensure judicial coherence and consistency across the judgment, the global context $\mathcal{G}$ must be referenced through the context adjustment function $q(\cdot)$:
\vspace{-8pt}
{
\setlength{\belowdisplayskip}{3pt} 
\begin{equation}
\hat{s}_i^{(0)} = q(s'_i, \mathcal{G})
\end{equation}}
The complete sentence translation process can be formalized as:
\vspace{-8pt}
{
\setlength{\belowdisplayskip}{3pt} 
\begin{equation}
\hat{s}_i^{(0)} = f_{\text{Trans}}(s_i, \hat{L}_i, \mathcal{G})
\end{equation}}where $\mathcal{G}$ represents the global context derived from preceding case facts and procedural history. This dual-agent approach ensures both lexical precision and syntactic fluidity, with the $\mathcal{A}{_\text{Trans}}$ dynamically adjusting translations based on feedback from the subsequent Expert Review Module through the iterative mechanism described in Equation~\ref{eq:iterative_update}.

\begin{table*}[htbp]
\centering
\caption{\small The performance (xCOMET-XL, \%) of various LLMs serving as agents in the Translation Execution Module and Expert Review Module. Best results in each column are in \textbf{bold}. Model abbreviations: 13B-C: Baichuan-13B-Chat, 13B-B: Baichuan-13B-Base, 3-6B: ChatGLM3-6B, 2-6B: ChatGLM2-6B. Results here use GPT-4o as the Translation Command Module agent; comparative results for GPT-4 and ChatGPT are detailed in Appendix~\ref{app:add}.}
\renewcommand{\arraystretch}{1.1} 
\setlength{\tabcolsep}{1.8pt}

\setlength{\aboverulesep}{1pt} 
\setlength{\belowrulesep}{1pt}

\setlength{\dashlinedash}{2pt} 
\setlength{\dashlinegap}{2pt}  

\resizebox{\textwidth}{!}{%
\tiny 
\begin{tabular}{ll ccc cc cc cc cc cc}
\toprule 
 & & \multicolumn{13}{c}{\textbf{Translation Command Module (GPT-4o)}} \\
\cmidrule(lr){3-15} 

\multicolumn{2}{l}{\textbf{Expert Review Module} $\downarrow$} & \multicolumn{13}{c}{\textbf{Translation Execution Module $\rightarrow$}} \\
\cmidrule(r){1-2} \cmidrule(lr){3-15}

 & & \multicolumn{3}{c}{OpenAI} 
 & \multicolumn{2}{c}{DeepSeek} 
 & \multicolumn{2}{c}{Qwen} 
 & \multicolumn{2}{c}{Baichuan} 
 & \multicolumn{2}{c}{ChatGLM} 
 & \multicolumn{2}{c}{ChatLaw} \\
\cmidrule(lr){3-5} \cmidrule(lr){6-7} \cmidrule(lr){8-9} \cmidrule(lr){10-11} \cmidrule(lr){12-13} \cmidrule(lr){14-15}

\textbf{Series} & \textbf{Model} & 
GPT-4o & GPT-4 & ChatGPT & 
MoE-16B & R1-Distill-32B & 
14B & 7B & 
13B-C & 13B-B & 
3-6B & 2-6B & 
33B & 13B \\
\midrule 

\multirow{3}{*}{OpenAI} 
 & \cellcolor{myblue}GPT-4o        & \cellcolor{myblue}\textbf{85.12} & \cellcolor{myblue}\textbf{84.87} & \cellcolor{myblue}\textbf{83.43} & \cellcolor{myblue}\textbf{84.19} & \cellcolor{myblue}\textbf{83.92} & \cellcolor{myblue}\textbf{83.08} & \cellcolor{myblue}\textbf{81.76} & \cellcolor{myblue}\textbf{82.34} & \cellcolor{myblue}\textbf{80.91} & \cellcolor{myblue}\textbf{81.47} & \cellcolor{myblue}\textbf{79.83} & \cellcolor{myblue}\textbf{80.52} & \cellcolor{myblue}\textbf{78.89} \\
 & GPT-4         & 84.88 & 84.24 & 83.17 & 83.91 & 83.68 & 82.79 & 81.42 & 82.03 & 80.57 & 81.18 & 79.44 & 80.16 & 78.43 \\
 & ChatGPT       & 84.13 & 83.56 & 82.29 & 82.94 & 82.71 & 81.83 & 80.77 & 81.15 & 79.88 & 80.42 & 78.76 & 79.37 & 77.82 \\
\cdashline{1-15}

\multirow{2}{*}{DeepSeek} 
 & MoE-16B            & 84.54 & 83.92 & 82.81 & 83.53 & 83.34 & 82.37 & 81.08 & 81.59 & 80.23 & 80.88 & 79.12 & 79.74 & 78.06 \\
 & R1-Distill-32B            & 84.27 & 83.65 & 82.58 & 83.41 & 83.25 & 82.16 & 80.93 & 81.44 & 80.09 & 80.71 & 78.98 & 79.62 & 77.91 \\
\cdashline{1-15}

\multirow{2}{*}{Qwen} 
 & 14B-Chat      & 83.76 & 83.18 & 82.04 & 82.87 & 82.59 & 81.86 & 80.62 & 81.03 & 79.67 & 80.35 & 78.54 & 79.18 & 77.45 \\
 & 7B-Chat       & 83.05 & 82.47 & 81.38 & 82.14 & 81.82 & 81.09 & 80.54 & 80.36 & 79.05 & 79.73 & 77.92 & 78.55 & 76.88 \\
\cdashline{1-15}

\multirow{2}{*}{Baichuan} 
 & 13B-Chat      & 83.38 & 82.73 & 81.65 & 82.42 & 82.16 & 81.33 & 80.27 & 81.33 & 79.36 & 80.08 & 78.25 & 78.84 & 77.19 \\
 & 13B-Base      & 82.44 & 81.88 & 80.79 & 81.56 & 81.23 & 80.47 & 79.38 & 79.94 & 79.60 & 79.15 & 77.37 & 77.93 & 76.24 \\
\cdashline{1-15}

\multirow{2}{*}{ChatGLM} 
 & 3-6B          & 83.02 & 82.35 & 81.27 & 82.05 & 81.76 & 80.94 & 79.82 & 80.44 & 79.18 & 80.27 & 77.85 & 78.46 & 76.73 \\
 & 2-6B          & 81.79 & 81.14 & 80.03 & 80.87 & 80.55 & 79.76 & 78.64 & 79.28 & 77.95 & 78.92 & 78.11 & 77.24 & 75.58 \\
\cdashline{1-15}

\multirow{2}{*}{ChatLaw} 
 & 33B           & 82.15 & 81.56 & 80.42 & 81.23 & 80.97 & 80.18 & 79.06 & 79.65 & 78.34 & 79.33 & 77.58 & 79.29 & 76.02 \\
 & 13B           & 80.46 & 79.88 & 78.74 & 79.52 & 79.25 & 78.43 & 77.36 & 77.97 & 76.65 & 77.62 & 75.83 & 77.41 & 76.26 \\

\bottomrule 
\end{tabular}%
}

\label{tab:2re}
\vspace{-0.3cm}
\end{table*}

\paragraph{Expert Review Module} As shown in Fig. \ref{fig:2}, $\mathcal{M}_{\text{Rev}}$ implements a comprehensive quality assurance system through four review agents. This module transforms initial translations into polished, authoritative outputs through systematic multi-dimensional evaluation and iterative refinement. The module employs four complementary review perspectives:
\begin{itemize}
\item The $\mathcal{A}_{\text{Align}}$ evaluates semantic accuracy by checking the precise alignment between source and target texts at both logical and factual levels consistency:
\vspace{-8pt}
{
\setlength{\belowdisplayskip}{3pt} \begin{equation}\delta^{(k)}_{i,\text{Align}} = f_{\text{Align}}(\hat{s}^{(k)}i, s_i, \mathcal{G}) \label{eq:align}\end{equation}}

\item $\mathcal{A}_{\text{TermR}}$ verifies legal term expressions, possessing the same access capability to HK legal glossary database $D$ as the $\mathcal{A}_{\text{Term}}$:
\vspace{-8pt}
{
\setlength{\belowdisplayskip}{3pt} 
\begin{equation} \delta^{(k)}_{i,\text{TermR}} = f_{\text{TermR}}(\hat{s}^{(k)}_i, \hat{L}_i, D) \label{eq:term_review} \end{equation}}
\item The $\mathcal{A}_{\text{Cita}}$ validates legal authorities, ensuring that all case references\footnote{Case references denote citations to legal precedents, e.g., \textit{HKSAR v. Chan Ka Ming} [2025] HKCFA 8.} and legislative provisions\footnote{Legislative provisions refer to specific statutory sections, e.g., Section 24 of the \textit{Safeguarding National Security Ordinance} (Cap.~A305).} adhere to the rigid HKCFA formatting standards:
\vspace{-8pt}
{
\setlength{\belowdisplayskip}{3pt} 
\begin{equation}
\delta^{(k)}_{i,\text{Cita}} = f_{\text{Cita}}(\hat{s}^{(k)}_i) \label{eq:cita}
\end{equation}}

\item The $\mathcal{A}_{\text{StyleP}}$ guarantees stylistic fidelity via polishing, ensuring the translation exhibits a formal judicial tone and syntactic fluency:
\vspace{-8pt}
{
\setlength{\belowdisplayskip}{3pt} 
\begin{equation}\delta^{(k)}_{i,\text{Style}} = f_{\text{Style}}(\hat{s}^{(k)}_i) \label{eq:style}\end{equation}}

\end{itemize}

These agents operate in concert within each review round, generating comprehensive feedback:
\vspace{-8pt}
{
\setlength{\belowdisplayskip}{3pt}
\begin{equation}\delta^{(k)}_i = \{\delta^{(k)}_{i,\text{Align}}, \delta^{(k)}_{i,\text{TermR}}, \delta^{(k)}_{i,\text{Cita}}, \delta^{(k)}_{i,\text{Style}}\} \label{eq:feedback_union}\end{equation}

When $\delta^{(k)}_i = \emptyset$, the translation achieves optimal quality and proceeds to final output. Otherwise, the feedback triggers another iteration of refinement through the Translation Execution Module. This multi-round review mechanism ensures that each translation meets the highest standards of semantic accuracy, terminological precision, citation compliance, and stylistic integrity, while preserving the rigorous nature of the source English judgment. The review mechanism maximally ensures the quality of the translation while reducing potential biases that might arise from errors within a single module. Detailed prompt templates for the seven agents are provided in Appendix~\ref{app:p}.

\vspace{-10pt}
\section{Evaluation}

This section presents both automated and human evaluations of TransLaw’s translation performance.

\vspace{-5pt}
\subsection{Automated evaluation}

Automated evaluation of an MT system needs to be conducted by using available automated evaluation metrics to estimate the quality scores of its translation outputs by comparing them with a given bilingual text dataset providing the gold standard answers.
\vspace{-10pt}
\paragraph{Metrics}
The metrics adopted for the evaluation are two of the most popular automated MT evaluation metrics in recent years: (1) xCOMET-XL, a version of xCOMET, which is a state-of-the-art learned metric for various levels of evaluation \cite{guerreiro2024xcomet}; and (2) wmt22-unite-da, a unified MT quality evaluation model \cite{guttmann2024comet}. To ensure statistical reliability, for all metrics, we report two times standard deviation using 1,000 runs of bootstrap \cite{efron1986bootstrap} on the test dataset, which corresponds to a 95.45\% confidence level under the assumption of a normal distribution.

\vspace{-10pt}
\paragraph{Evaluated LLM Models}
We evaluated 13 widely adopted LLMs, classified into two categories: General and Legal-specific LLMs. For General LLMs, the testbed includes GPT-4o \cite{hurst2024gpt}, GPT-4 \cite{achiam2023gpt}, ChatGPT \cite{ouyang2022training}, two models from DeepSeek (DeepSeekMoE-16B-Base, a 16B mixture-of-experts model from \citet{dai2024deepseekmoe}, and DeepSeek-R1-Distill-Qwen-32B, a 32B distilled variant of DeepSeek-R1 reported in \citet{guo2025deepseekr1}), the ChatGLM series (ChatGLM2-6B, ChatGLM3-6B) \cite{zeng2023glm}, the Baichuan series (Baichuan-13B-Base, Baichuan-13B-Chat) \cite{yang2023baichuan}, and the Qwen series (Qwen-7B-Chat, Qwen-14B-Chat) \cite{bai2023qwen}; for Legal-specific LLMs, we included ChatLaw-13B and ChatLaw-33B \cite{cui2023chatlaw}. Detailed links to the LLM models are provided in Appendix~\ref{app:exp}. TransLaw consists of three modules. For the agents in each module, we employ the same LLM from the list above.

\paragraph{Analysis of Evaluation Results}

   The experimental results are shown in Tables~\ref{tab:1re} and~\ref{tab:2re}. These results reveal the following points.
\vspace{-10pt}
\begin{itemize}
   \item All open-source LLMs perform slightly under the closed-source (commercial) ones like GPT-4o and GPT-4, which achieve  the best performance in this benchmark. However, due to insufficient knowledge of Hong Kong's legal system, both they exhibit substantial limitations in legal judgment translation. This indicates significant potential for improvement in legal-domain LLM capabilities.
   \vspace{-10pt}
   \item Increased LLM model scale consistently enhances performance. For instance, Qwen-14B outperforms Qwen-7B. Furthermore, chat-optimized LLMs (e.g., Baichuan-13B-Chat) surpass their base models (e.g., Baichuan-13B-Base), suggesting that greater instruction-following capabilities, gained through supervised fine-tuning and alignment optimization, can be more effective in unlocking LLMs' potential in translation.
   \vspace{-10pt}
   \item Surprisingly, legal-specific LLMs do not always outperform general LLMs. We speculate that there are two possible reasons. First, the capability of these legal-specific LLMs could be limited by their base models, which are usually known not to be as strong as other LLMs such as GPT-4o and GPT-4; moreover, the continuous pre-training or fine-tuning using legal corpora may not further promote the abilities of the original base models. It is also possible that both play a part in explaining this result, which certainly suggests the necessity for further design to improve the performance of legal LLMs.

\end{itemize}

Contrasting the TransLaw performance in Columns 6-8 of Table~\ref{tab:1re} with the Single Translator Agent reveals marked performance improvements, which demonstrate the efficacy of the collaborative TransLaw framework. Despite these inevitable limitations of TransLaw, the above evaluation shows that using LLMs as role-specific agents in a MAS can effectively assist translation tasks to facilitate the Hong Kong bilingual legal system. However, investigating how LLMs compare with humans taking similar translation roles, like annotation and proofreading, paves the way for powerful intelligent legal LLMs that improve the efficiency and quality of legal translation services.

\paragraph{Ablation Study}
As shown in Table~\ref{tab:ablation}, we conduct ablation studies
across 3 LLMs for each component, ablating the RAG HK legal glossary
database ($\mathcal{D}$), the context memory ($G$), and each of the
seven agents in $\mathcal{C}$. We report xCOMET-XL together with the
number of iterations required to reach the termination condition
$\delta_i^{(k)}=\emptyset$. The results indicate that every component
contributes a significant performance gain, but the contributions are
sharply unequal. Removing the Sentence Translation Agent or the
Translation Command Agent collapses the system, yielding the two
largest drops. The Legal Terminology Agent and the RAG database are
tightly coupled and rank next, confirming that LLMs' parametric knowledge
cannot substitute for an authoritative domain glossary. The context
memory consistently degrades performance when removed, confirming that
sentence-level translation is insufficient for document-level coherence.
Within the Expert Review Module, removing any of the Semantic Alignment,
Legal Term Review, or Legal Citation agents leads to a clear decline,
while removing the Stylistic Polishing agent yields a smaller yet
consistent decrease.

\paragraph{Iteration Efficiency}
As shown in Table~\ref{tab:ablation}, the maximum number of iterations
required under TransLaw is bounded to 3 across evaluated LLMs, showing
that the multi-agent feedback is highly actionable and effectively
prevents infinite loops. Stronger LLMs converge faster than weaker open
models, and 2 to 3 iterations emerge as the efficient operating point
for expert-level legal translation.

\begin{table}[h]
\centering
\caption{Ablation results on three LLMs. ``Score'' denotes xCOMET-XL
(\%); ``Iter'' denotes the number of iterations to reach the termination
condition. $\mathcal{D}$ denotes the RAG HK legal glossary database;
$G$ denotes the context memory; $\mathcal{A}_{\text{Com}}$,
$\mathcal{A}_{\text{Term}}$, and $\mathcal{A}_{\text{Trans}}$ denote
the Translation Command, Legal Terminology, and Sentence Translation
agents, respectively; $\mathcal{A}_{\text{Align}}$,
$\mathcal{A}_{\text{TermR}}$, $\mathcal{A}_{\text{Cita}}$, and
$\mathcal{A}_{\text{StyleP}}$ denote the Semantic Alignment, Legal
Term Review, Legal Citation, and Stylistic Polishing agents,
respectively.}
\small
\setlength{\tabcolsep}{5pt}
\renewcommand{\arraystretch}{1.15}
\setlength{\heavyrulewidth}{1.2pt}
\setlength{\lightrulewidth}{1.2pt}
\begin{tabular}{lcccccc}
\toprule
\multirow{2}{*}{\textbf{Setting}}
 & \multicolumn{2}{c}{\textbf{Qwen-14B}}
 & \multicolumn{2}{c}{\textbf{Baichuan-13B-C}}
 & \multicolumn{2}{c}{\textbf{ChatLaw-33B}} \\
\cmidrule(lr){2-3} \cmidrule(lr){4-5} \cmidrule(lr){6-7}
 & Score & Iter & Score & Iter & Score & Iter \\
\midrule
\rowcolor{myblue}
TransLaw                          & \textbf{81.86} & 2 & \textbf{81.33} & 3 & \textbf{79.29} & 2 \\
w/o $\mathcal{D}$                 & 77.11 & -- & 76.72 & -- & 74.57 & -- \\
w/o $G$                           & 79.38 & -- & 78.77 & -- & 76.65 & -- \\
\midrule
w/o $\mathcal{A}_{\text{Com}}$    & 74.85 & -- & 74.32 & -- & 72.18 & -- \\
w/o $\mathcal{A}_{\text{Term}}$   & 77.94 & 2 & 77.46 & 3 & 75.21 & 2 \\
w/o $\mathcal{A}_{\text{Trans}}$  & 68.42 & -- & 67.89 & -- & 65.74 & -- \\
\midrule
w/o $\mathcal{A}_{\text{Align}}$  & 78.81 & 2 & 78.36 & 2 & 76.20 & 2 \\
w/o $\mathcal{A}_{\text{TermR}}$  & 79.52 & 2 & 78.89 & 2 & 76.82 & 2 \\
w/o $\mathcal{A}_{\text{Cita}}$   & 79.86 & 2 & 79.24 & 2 & 77.39 & 2 \\
w/o $\mathcal{A}_{\text{StyleP}}$ & 80.94 & 2 & 80.49 & 3 & 78.45 & 2 \\
\bottomrule
\end{tabular}

\label{tab:ablation}
\end{table}

\subsection{Human Evaluation}
For human evaluation, a scoring scheme needs to be formulated to integrate human evaluation scores in various evaluation dimensions into one. Therefore, we propose the legal ACS metric for the translation of HK legal judgments, which consists of three dimensions: \textbf{A} (accuracy of legal meaning), \textbf{C} (coherence and cohesion in structure), and \textbf{S} (style appropriateness) (henceforth ACS), whose formulation is presented below, followed by the settings and results of our human evaluation.

\paragraph{Test Set}

We selected the bilingual texts of the judgment ``HKSAR - Court of Final Appeal - Final Appeal Criminal Case No.~1 of 2021'' (henceforth FACC~1/2021 for brevity; see Appendix~\ref{app:judgment_sample}) from the HKCFA Judgment 97-22 Dataset as our human evaluation data. Following paragraph-level segmentation and manual alignment, the whole test set consists of 200 paragraph-level source-target pairs. This human evaluation set comprises 12,029 tokens in English and 19,478 tokens in Chinese.
\vspace{-10pt}
\paragraph{Evaluation Metrics}
Aiming at a comprehensive, adequate and reliable evaluation of the translation quality of HK legal judgments, the ACS metric is formulated as $I = \alpha A + \beta C + \gamma S$, where $A$, $C$, and $S$ are the scores given by human expert evaluators in the three key dimensions, and $\alpha$, $\beta$, and $\gamma$ are respective weight coefficients according to their relative importance. Based on the experience and recommendation of domain experts, these weights are set as follows for our manual evaluation of legal judgment translation: \(\alpha = 0.6, \beta = 0.3, \gamma = 0.1\). This setting recognizes the most fundamental role of accuracy. In Figure \ref{fig:human_eval_full}, we further set different weights for evaluation, which remains an interesting issue for further examination, as is the human evaluation of TransLaw's performance; both are expected to give meaningful hints to justify this scoring scheme.

\vspace{-10pt}
\paragraph{Setup}
We perform a manual evaluation of the FACC 1/2021 test set, comparing the official human translation against two GPT-4o system configurations (the best-performing model in Table \ref{tab:1re}): one operating as a Single Translator Agent and the other deployed within the TransLaw framework. To mitigate human evaluator fatigue, segment lengths were controlled, with the maximum length reaching 234 English words (290 tokens) and 414 Chinese words (580 tokens). Anonymized evaluation tables containing segment/sentence IDs, source texts, and system outputs were assessed by 10 certified professional legal translators using a 0-10 point scale across three predefined dimensions. The evaluation guidelines for human experts can be found in Appendix~\ref{app:guide} (Figure~\ref{fig:annotation_guidelines}).

\begin{figure}[t!] 
    \centering

    \begin{subfigure}{\linewidth} 
        \centering
        \includegraphics[width=\linewidth]{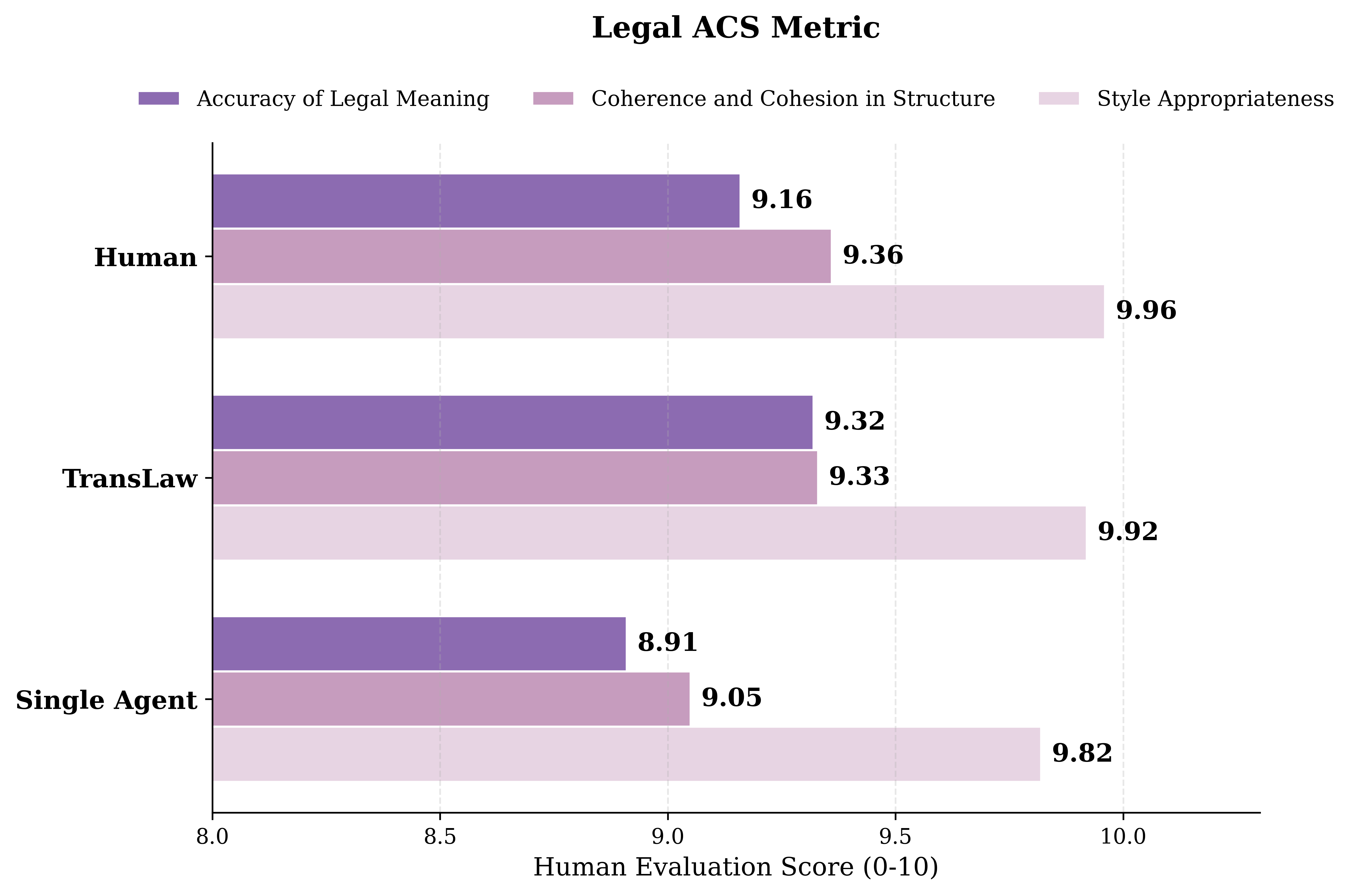} 
        \label{fig:detailed_metrics}
    \end{subfigure}
    
    \vspace{0.1cm} 
    
    \begin{subfigure}{\linewidth}
        \centering
        \includegraphics[width=\linewidth]{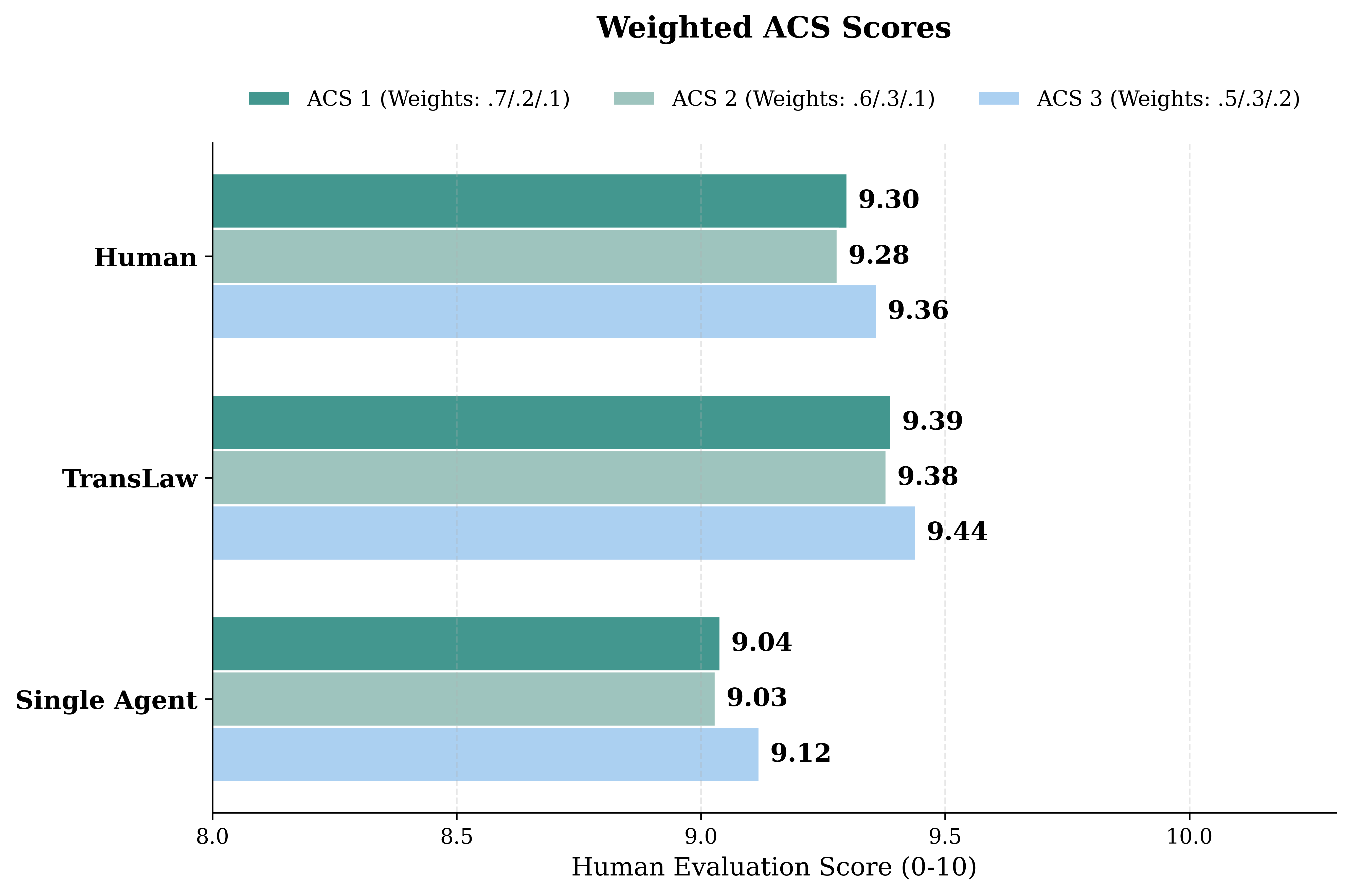}
        \label{fig:weighted_scores}
    \end{subfigure}
    
    \caption{Human Evaluation Results. Performance of the three systems across three dimensions (top) and different weighting schemes (bottom).}
    \label{fig:human_eval_full}
    \vspace{-0.3cm}
\end{figure}

\paragraph{Results and Analysis}

Figure \ref{fig:human_eval_full} presents human evaluation scores for the three configurations using ACS metrics. TransLaw achieves superior Accuracy in conveying legal meanings and leads all ACS metrics across three different weighting schemes. Official human translation demonstrates relative strengths in Coherence and Cohesion as well as Style, indicating that human translation capability remains superior to TransLaw. The three different weighting schemes for ACS in Figure \ref{fig:human_eval_full} yielded very minor differences in ACS scores, convincingly justifying the soundness and consistency of this scoring scheme for human evaluation. More detailed human evaluation of translation cases is provided in Appendix~\ref{app:Cases_study}.

\section{Cost Analysis}
\label{sec:cost_analysis}
The cost of human translation can vary based on several factors, including the type of text, the translator's location, and their level of experience. The American Translators Association recommends a minimum charge of US\$0.12 per word for professional translation. Accordingly, translating a judgment like FACC 1/2021, totaling 11,585 words (12,029 tokens), would cost US\$1,390.20. In contrast, the cost of translating the FACC 1/2021 test set using GPT-4o is approximately US\$0.39. Using TransLaw, the cost breaks down to approximately US\$0.35. This contrast indicates that using TransLaw to translate HK legal judgments can reduce translation costs by roughly three orders of magnitude compared to human translation and by 10.26\% compared to GPT-4o. Note that US\$0.39 and US\$0.35 for using GPT-4o and TransLaw are API (Application Programming Interface) costs, excluding the cost for human proofreading and editing of the translation output from an API. Given that the Avg. standard rate for human editing is approximately US\$0.04 per word\footnote{\url{https://www.translationedge.com/pricing}.}, the total cost for translating plus editing the said judgment would be US\$0.35 + US\$0.04 $\times$ 11,585 = US\$463.75, saving US\$926.45, or two-thirds of the full human translation cost.

\section{Conclusion}

In this paper, we constructed and released HKCFA Judgment 97-22, a large-scale bilingual dataset, and conducted the first comprehensive evaluation of LLM capabilities in Hong Kong case law translation. To address the limitations of current models, we proposed TransLaw, a collaborative Multi-Agent System (MAS) designed to mimic professional human translation workflows. Our experimental results confirm the validity and effectiveness of these collaborative strategies, while also providing a comparative benchmark of various LLMs acting as agents. Overall, TransLaw proves to be a robust framework for handling the complexities of Hong Kong case law translation, laying a foundation for future research in this area. In future work, we aim to integrate human expert feedback to refine initial reasoning steps, thereby further enhancing translation accuracy.

\section{Ethics Statement}
Our dataset and evaluation benchmark contain no personal, sensitive, or private information; they consist solely of publicly available data.

\section*{Impact Statement}

This paper presents work whose goal is to advance the field of NLP for Social Good, specifically in English-to-Chinese Hong Kong case law translation. Our work fulfills a constitutional obligation under Articles 8 and 9 of the Basic Law, serves as an essential mechanism ensuring jurisprudential precision and facilitating cross-jurisdictional legal communication, and addresses a scalability crisis manual workflows cannot sustain. We hope our work offers practical value by helping to reduce language barriers for the general public, alleviating the scalability challenges in judicial translation, and supporting Hong Kong's constitutional bilingual legal framework. There are many potential societal consequences of our work, none which we feel must be specifically highlighted here.

\section*{acknowledgement}
The research reported in this paper was supported
by the Research Grants Council (RGC) of HKSAR,
China through its FDS grant UGC/FDS51/H03/20
and GRF grant 11602524.

\nocite{langley00}

\bibliography{example_paper}
\bibliographystyle{icml2026}

\newpage
\appendix
\onecolumn
\appendix

\clearpage

\twocolumn 
\onecolumn

\section{Use of AI Assistants}
We used Claude Opus 4.7 and Sonnet 4.6 for coding, shortening texts and editing LaTeX more efficiently.

\section{HKACFA Judgment Example (Case No. FACC 1/2021)}
\label{app:judgment_sample}

\begin{figure}[H] 
    \centering
    
    \setlength{\fboxrule}{1pt} 
    \setlength{\fboxsep}{0pt}   

    \begin{subfigure}[b]{0.48\textwidth}
        \centering
        \fbox{\includegraphics[page=1, width=0.98\linewidth]{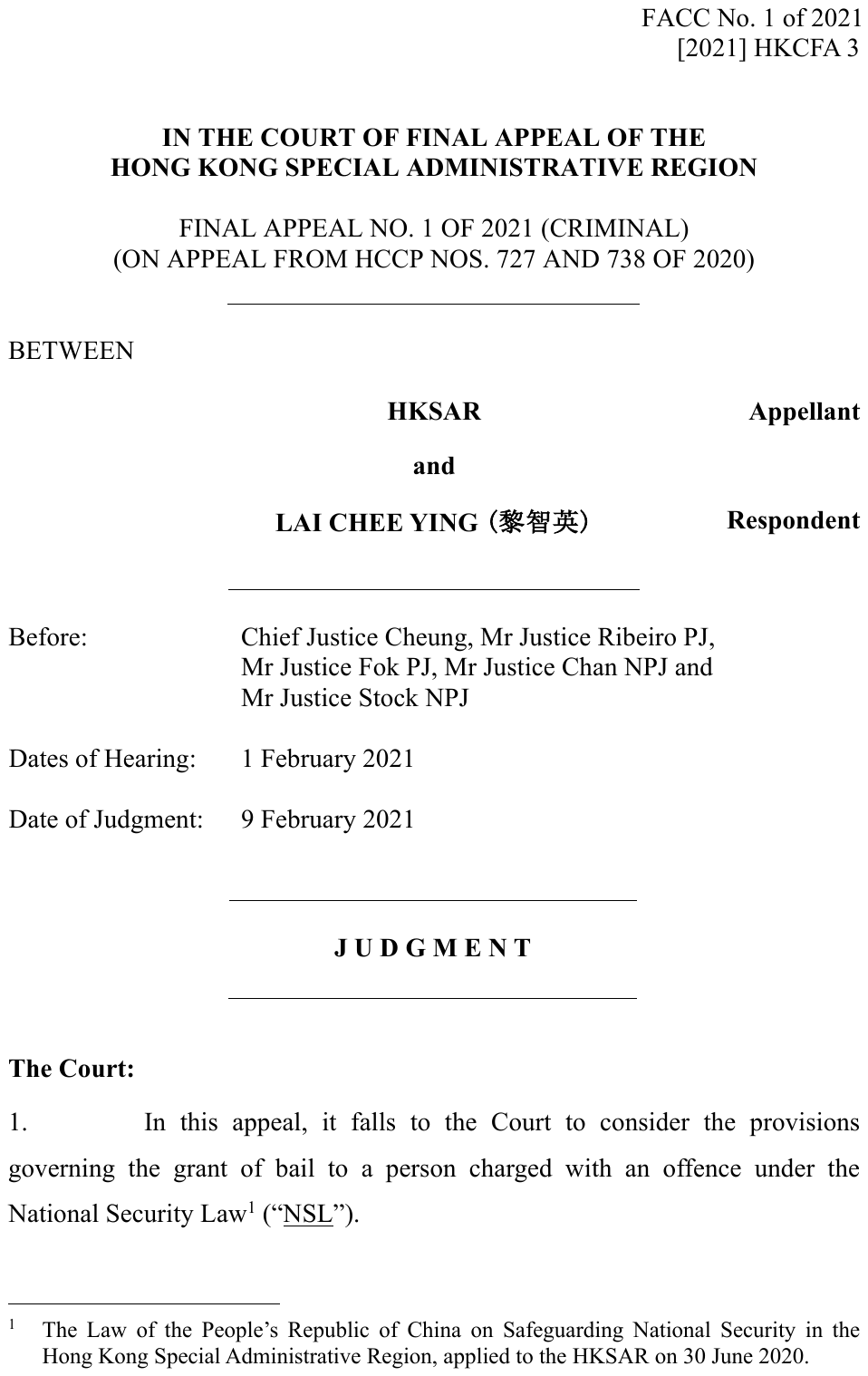}}
        \caption{English Judgment (Source)}
        \label{fig:eng_judgment}
    \end{subfigure}
    \hfill 
    \begin{subfigure}[b]{0.48\textwidth}
        \centering
        \fbox{\includegraphics[page=1, width=0.98\linewidth]{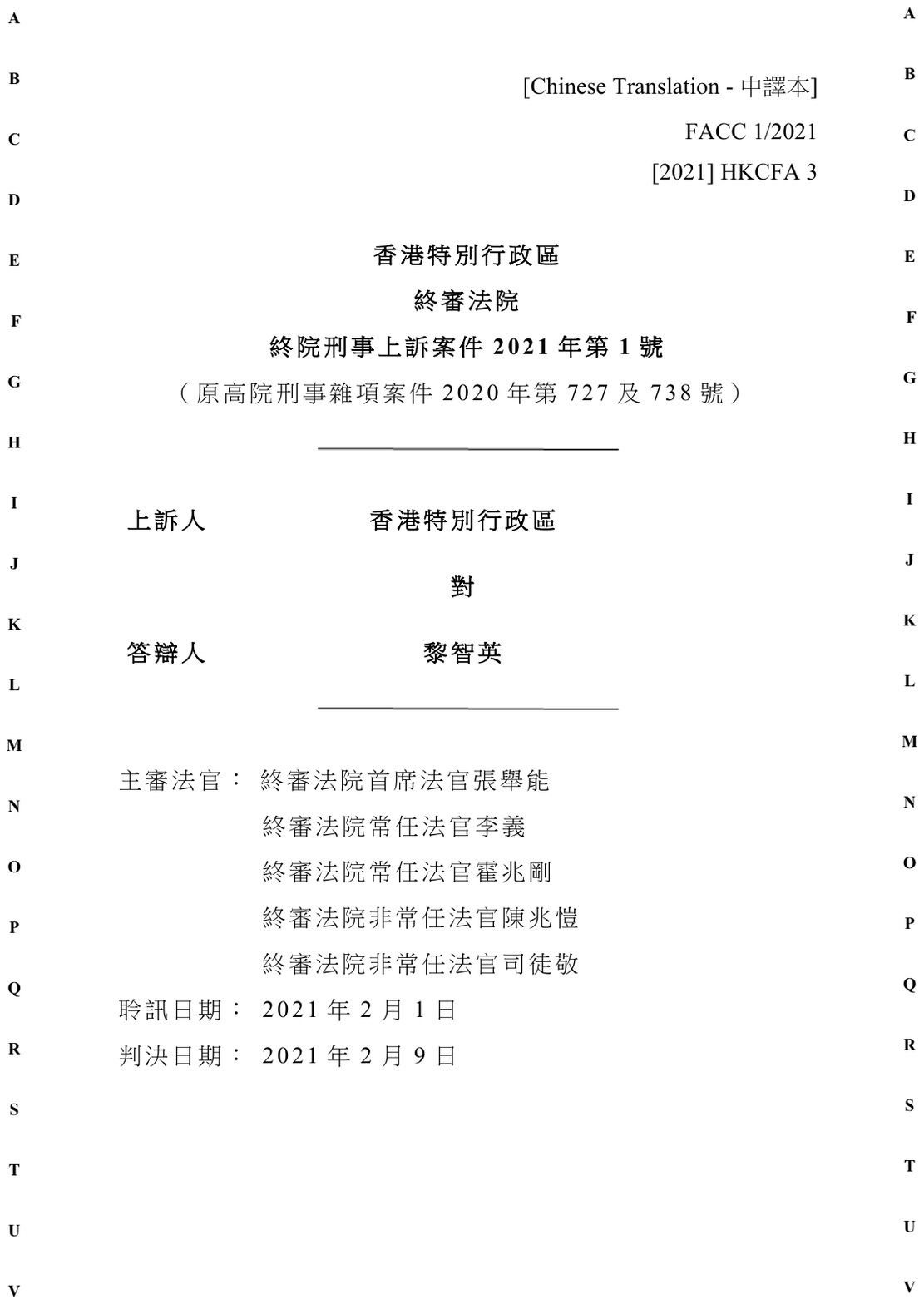}}
        \caption{Chinese Translation (Target)}
        \label{fig:chn_judgment}
    \end{subfigure}
    
    \caption{Example of the first page of a bilingual HKCFA judgment (Case No. FACC 1/2021), comparing the source text and target translation.}
    \label{fig:judgment_comparison}
\end{figure}

\clearpage

\section{Evaluation Codes}
\label{app:Proofread_code}

\begin{table*}[h]
\centering
\caption{Evaluation Codes}
\small
\begin{tabular}{p{0.20\textwidth} p{0.15\textwidth} p{0.6\textwidth}}
\Xhline{1.4pt} 
\textbf{Evaluation Dimension} & \textbf{Sub-dimension} & \textbf{Description} \\
\hline 

\multirow{10}{*}{\textbf{Accuracy}} 
 & CW & Choice of word. The word or expression is not a good choice. \\
 & IF & Information structure not preserved. \\
 & MC & Meaning has been changed because of inappropriate restructuring. \\
 & MT & Mistranslation due to inadequate comprehension or misinterpretation. \\
 & NA & The translation conveys a different meaning from that of the source text. \\
 & NC & Meaning not clear due to ambiguity, vagueness or syntactic problems. \\
 & OM & Omission. Part of the original has been left untranslated. \\
 & OT & Over-translation. Too much has been read into the source text. \\
 & TL & Too literal, affecting comprehensibility. \\
 & UT & Under-translation. Meaning is not adequately captured in translation. \\
\hline 

\multirow{12}{*}{\textbf{Grammar}} 
 & Art & Article. \\
 & Det & Determiner. \\
 & MD & Modality. \\
 & NB & Number. \\
 & PN & Punctuation. \\
 & Prep & Wrong preposition. \\
 & PS & Part of speech. \\
 & SP & Spelling or wrong character. \\
 & ST & The sentence or part of the sentence is ill-formed or ambiguous. \\
 & SV & Subject verb agreement. \\
 & TN & Tense problem. \\
 & WO & Word order. \\
\hline 

\multirow{9}{*}{\textbf{Usage and style}} 
 & CL & Collocation problem. \\
 & CN & The word or expression has connotation not appropriate in the context. \\
 & CO & Connective problem, e.g., inappropriate connectives. \\
 & IC & Inconsistent use of a word; or incoherence between clauses or sentences. \\
 & ID & Idiomaticity, i.e., unidiomatic expression. \\
 & RF & Reference problem, e.g., ambiguous use of a pronoun. \\
 & RN & Redundancy: the word or expression should be deleted. \\
 & SL & Stylistic problems, e.g., the word or expression is not of an appropriate style. \\
 & TS & Transition problems: sentences not well connected; bad language flow. \\
\Xhline{1.4pt} 
\end{tabular}
\end{table*}
\clearpage

\section{More Cases}
\label{app:Cases_study}

\begin{table*}[h]
    \centering
    \small
     \caption{Case Study: Comparative analysis of the Source Text translation using GPT-4o acting as a single translator agent and TransLaw, alongside the Reference Text.}
    \begin{CJK*}{UTF8}{bsmi}
    \begin{tabular}{@{}>{\raggedright}p{1.5cm} p{14cm}@{}}
        \Xhline{1.4pt}
        Source Text & At present, the increasingly notable national security risks in the HKSAR have become a prominent problem. In particular, since the onset of HK's 'legislative amendment turmoil' in 2019, anti-China forces seeking to disrupt HK have blatantly advocated such notions as 'HK independence', 'self-determination' and 'referendum', and engaged in activities to undermine national unity and split the country. They have brazenly desecrated and defiled the national flag and emblem, incited HK people to oppose China and the Communist Party of China ('CPC'), besiege Central People's Government ('CPG') offices in HK, and discriminate and ostracize Mainland personnel in HK. These forces have also wilfully disrupted social order in HK, violently resisted police enforcement of the law, damaged public facilities and property, and paralyzed governance by the government and operation of the legislature. Moreover, in recent years, certain foreign or external forces have flagrantly interfered in HK's affairs. They have made intervention and created disturbances in various ways, such as by legislative and administrative means and through non-governmental organizations. In collusion with those anti-China HK disrupters, these forces of the same ilk backed and cheered on the disrupters and provided a protective umbrella, and utilized HK to carry out activities endangering national security. These acts and activities have seriously challenged the bottom line of the 'One Country, Two Systems' principle, seriously undermined the rule of law, and seriously jeopardized national sovereignty, security and development interests. \\ 
        \midrule
        GPT-4o & 目前，香港特別行政區日益突出 \textcolor{lightred}{[CW]} 的國家安全風險已成為一個顯著 \textcolor{lightred}{[CW]} 問題。特別是自2019年香港的「立法修訂風波」開始 \textcolor{lightred}{[ST]}，反華 \textcolor{lightred}{[NA]} 勢力企圖破壞香港的行為愈發公然，公然宣揚如「香港獨立」、「自決」和「公投」等概念，並從事破壞國家統一和分裂國家的活動。他們 [Pronoun] 公然褻瀆和污損國旗及國徽，煽動香港人反對中國及中國共產黨（「中共」），圍攻中央人民政府（「中央政府」）在香港的辦事處，歧視和排斥內地人員。這些勢力還任意破壞香港的社會秩序，暴力抗拒警方執法，破壞公共設施和財產，使政府治理 \textcolor{lightred}{[CW]} 和立法機構運作陷於癱瘓。\textcolor{lightred}{[OM]} 近年來，某些外國或外部 \textcolor{lightred}{[CW]} 勢力公然干預香港事務，以立法和行政手段及 \textcolor{lightred}{[US]} 透過非政府組織的各種方式 \textcolor{lightred}{[NC]} 進行干擾 \textcolor{lightred}{[CL]} 和製造混亂。這些勢力 \textcolor{lightred}{[UT]} 與反華香港破壞分子 \textcolor{lightred}{[CW/NC]} 相勾結，支持並鼓舞破壞者，提供保護傘，利用香港從事危害國家安全的活動。這些行為和活動嚴重挑戰了「一國兩制」原則的底線，嚴重破壞了法治，嚴重危害了國家主權、安全和發展利益。 \\ 
        \midrule
        TransLaw & 目前，香港特別行政區日益突顯的國家安全風險已成為一個突出問題。特別是自2019年香港「修例風波」爆發 \textcolor{lightred}{[OT]} 以來，一些 \textcolor{lightred}{[NA]} 反中勢力企圖擾亂香港，公然宣揚「港獨」、「自決」及「公投」等概念 \textcolor{lightred}{[CW]}，並從事破壞國家統一和分裂國家的活動。他們 [Pronoun] 公然褻瀆和污損國旗及國徽，煽動香港人反對中國和中國共產黨（「中共」），圍攻中央人民政府（「中央政府」）在港機構，以及歧視和排斥香港的內地人員。這些勢力還肆意擾亂香港社會秩序，暴力抗拒警方執法，破壞公共設施和財產，並癱瘓政府治理和立法機構的運作。\textcolor{lightred}{[OM]} 近年來，一些 \textcolor{lightred}{[NA]} 外國或外部 \textcolor{lightred}{[CW]} 勢力公然干預香港事務，通過立法和行政手段以及非政府組織等多種方式進行干擾和創造混亂。這些勢力與反中香港破壞分子相勾結 \textcolor{lightred}{[NA]}，支持並為其助威 \textcolor{lightred}{[ST]}，提供保護傘，利用香港進行危害國家安全的活動。這些行為和活動嚴重挑戰了「一國兩制」原則的底線，嚴重破壞了法治，嚴重危害了國家主權、安全和發展利益。 \\ 
        \midrule
        Reference Text & 目前，香港特別行政區日益凸顯的國家安全風險已成為一個突出問題。特別是2019年香港發生「修例風波」以來，反中亂港勢力公然鼓吹「港獨」、「自決」、「公投」等概念，從事破壞國家統一、分裂國家的活動；公然侮辱、污損國旗國徽，煽動港人反中反共、圍攻中央駐港機構、歧視和排擠內地在港人員；蓄意破壞香港社會秩序，暴力對抗警方執法，毀損公共設施和財物，癱瘓政府管治和立法會運作。近年來，某些外國和境外勢力公然干預香港事務，通過立法、行政、非政府組織等多種方式進行干擾和和製造混亂，與香港反中亂港勢力勾連合流、沆瀣一氣，為香港反中亂港勢力撐腰打氣、提供保護傘，利用香港從事危害我國國家安全的活動。這些行為和活動，嚴重挑戰「一國兩制」原則底線，嚴重損害法治，嚴重危害了國家主權、安全和發展利益。\\ 
        \Xhline{1.4pt}
    \end{tabular}
   
    \end{CJK*}
    
    \label{tab:translation_comparison}
\end{table*}

\clearpage
\begin{CJK*}{UTF8}{bsmi}

\section{Overview of LLMs in the experiment}
\label{app:exp}

\begin{table*}[htbp]
\centering
\caption{Overview of the Large Language Models (LLMs) evaluated in our experiments.}
\begin{adjustbox}{width=\textwidth, totalheight=\textheight, keepaspectratio}
    \begin{tabular}{@{} l c c p{1cm} p{10.5cm} @{}}
\toprule[1.5pt]
\textbf{Model} & \textbf{Size} & \textbf{Seq\_len} & \textbf{Access} & \textbf{Url} \\
\midrule
GPT-4o               & N/A    & 8192      & API     & \url{https://platform.openai.com/docs/overview }     \\
\cline{1-5}
GPT-4          & N/A    & 8192   & API     & \url{https://platform.openai.com/docs/overview }     \\
\cline{1-5}
ChatGPT        & N/A    & 4096   & API     & \url{https://platform.openai.com/docs/overview }     \\
\cline{1-5}
Qwen-Chat-7B      & 7B     & 8192   & Weights & \url{https://huggingface.co/Qwen/Qwen-7B-Chat }     \\
\cline{1-5}
Qwen-Chat-14B      & 14B    & 8192   & Weights & \url{https://huggingface.co/Qwen/Qwen-14B-Chat }     \\
\cline{1-5}
DeepSeekMoE & 16B    & 4k     & Weights & \url{https://huggingface.co/deepseek-ai/deepseek-moe-16b-base}     \\
\cline{1-5}
DeepSeek-R1-Distill-Qwen & 32B    & 8192   & Weights & \url{https://huggingface.co/deepseek-ai/DeepSeek-R1-Distill-Qwen-32B}     \\
\cline{1-5}
ChatGLM-6B           & 6B     & 2048   & Weights & \url{https://huggingface.co/THUDM/chatglm-6b }     \\
\cline{1-5}
ChatGLM2-6B          & 6B     & 8192   & Weights & \url{https://huggingface.co/THUDM/chatglm2-6b }     \\
\cline{1-5}
ChatGLM3-6B          & 6B     & 8192   & Weights & \url{https://huggingface.co/THUDM/chatglm3-6b }     \\
\cline{1-5}
Baichuan-7B-Base     & 7B     & 4096   & Weights & \url{https://huggingface.co/baichuan-inc/Baichuan-7B}     \\
\cline{1-5}
Baichuan-13B-Base    & 13B    & 4096   & Weights & \url{https://huggingface.co/baichuan-inc/Baichuan-13B-Base}     \\
\cline{1-5}
Baichuan-13B-Chat    & 13B    & 4096   & Weights & \url{https://huggingface.co/baichuan-inc/Baichuan-13B-Chat}     \\
\cline{1-5}
ChatLaw-13B          & 13B    & 2048   & Weights & \url{https://huggingface.co/pandalla/ChatLaw-13B}     \\
\cline{1-5}
ChatLaw-33B          & 33B    & 2048   & Weights & \url{https://huggingface.co/pandalla/ChatLaw-33B}     \\
\midrule[1.5pt]
\end{tabular}
\end{adjustbox}
\label{tab:llm}
\end{table*}

\clearpage
\onecolumn
\section{Additional Experimental Results}
\label{app:add}
\begin{table*}[htbp]
\centering
\caption{\small The performance (xCOMET-XL, \%) using \textbf{GPT-4} as the Translation Command Module agent. Best results in each column are in \textbf{bold}. Model abbreviations: 13B-C: Baichuan-13B-Chat, 13B-B: Baichuan-13B-Base, 3-6B: ChatGLM3-6B, 2-6B: ChatGLM2-6B.}
\renewcommand{\arraystretch}{1.1} 
\setlength{\tabcolsep}{1.8pt}

\setlength{\aboverulesep}{1pt} 
\setlength{\belowrulesep}{1pt}

\setlength{\dashlinedash}{2pt} 
\setlength{\dashlinegap}{2pt}  

\resizebox{\textwidth}{!}{%
\tiny 
\begin{tabular}{ll ccc cc cc cc cc cc}
\toprule 
 & & \multicolumn{13}{c}{\textbf{Translation Command Module (GPT-4)}} \\
\cmidrule(lr){3-15} 

\multicolumn{2}{l}{\textbf{Expert Review Module} $\downarrow$} & \multicolumn{13}{c}{\textbf{Translation Execution Module $\rightarrow$}} \\
\cmidrule(r){1-2} \cmidrule(lr){3-15}

 & & \multicolumn{3}{c}{OpenAI} 
 & \multicolumn{2}{c}{DeepSeek} 
 & \multicolumn{2}{c}{Qwen} 
 & \multicolumn{2}{c}{Baichuan} 
 & \multicolumn{2}{c}{ChatGLM} 
 & \multicolumn{2}{c}{ChatLaw} \\
\cmidrule(lr){3-5} \cmidrule(lr){6-7} \cmidrule(lr){8-9} \cmidrule(lr){10-11} \cmidrule(lr){12-13} \cmidrule(lr){14-15}

\textbf{Series} & \textbf{Model} & 
GPT-4o & GPT-4 & ChatGPT & 
MoE-16B & R1-Distill-32B & 
14B & 7B & 
13B-C & 13B-B & 
3-6B & 2-6B & 
33B & 13B \\
\midrule 

\multirow{3}{*}{OpenAI} 
 & \cellcolor{myblue}GPT-4o        & \cellcolor{myblue}\textbf{84.78} & \cellcolor{myblue}\textbf{84.42} & \cellcolor{myblue}\textbf{83.11} & \cellcolor{myblue}\textbf{83.84} & \cellcolor{myblue}\textbf{83.56} & \cellcolor{myblue}\textbf{82.72} & \cellcolor{myblue}\textbf{81.43} & \cellcolor{myblue}\textbf{81.98} & \cellcolor{myblue}\textbf{80.64} & \cellcolor{myblue}\textbf{81.12} & \cellcolor{myblue}\textbf{79.49} & \cellcolor{myblue}\textbf{80.17} & \cellcolor{myblue}\textbf{78.52} \\
 & GPT-4         & 84.53 & 84.09 & 82.86 & 83.57 & 83.33 & 82.44 & 81.18 & 81.67 & 80.29 & 80.83 & 79.11 & 79.82 & 78.14 \\
 & ChatGPT       & 83.79 & 83.21 & 81.97 & 82.63 & 82.41 & 81.52 & 80.46 & 80.84 & 79.53 & 80.11 & 78.42 & 78.96 & 77.47 \\
\cdashline{1-15}

\multirow{2}{*}{DeepSeek} 
 & MoE-16B            & 84.18 & 83.57 & 82.49 & 83.22 & 82.98 & 82.03 & 80.74 & 81.26 & 79.89 & 80.54 & 78.78 & 79.41 & 77.73 \\
 & R1-Distill-32B            & 83.92 & 83.31 & 82.26 & 83.08 & 82.91 & 81.84 & 80.62 & 81.12 & 79.76 & 80.37 & 78.63 & 79.28 & 77.59 \\
\cdashline{1-15}

\multirow{2}{*}{Qwen} 
 & 14B-Chat      & 83.41 & 82.84 & 81.72 & 82.53 & 82.27 & 81.51 & 80.29 & 80.68 & 79.33 & 80.02 & 78.21 & 78.84 & 77.12 \\
 & 7B-Chat       & 82.72 & 82.13 & 81.06 & 81.81 & 81.49 & 80.76 & 80.18 & 80.03 & 78.71 & 79.38 & 77.57 & 78.23 & 76.54 \\
\cdashline{1-15}

\multirow{2}{*}{Baichuan} 
 & 13B-Chat      & 83.04 & 82.38 & 81.33 & 82.08 & 81.83 & 81.02 & 79.94 & 80.97 & 79.02 & 79.74 & 77.92 & 78.51 & 76.86 \\
 & 13B-Base      & 82.11 & 81.53 & 80.47 & 81.24 & 80.89 & 80.13 & 79.04 & 79.62 & 79.28 & 78.81 & 77.03 & 77.61 & 75.91 \\
\cdashline{1-15}

\multirow{2}{*}{ChatGLM} 
 & 3-6B          & 82.69 & 82.01 & 80.94 & 81.73 & 81.42 & 80.61 & 79.48 & 80.12 & 78.84 & 79.93 & 77.51 & 78.14 & 76.39 \\
 & 2-6B          & 81.46 & 80.82 & 79.71 & 80.54 & 80.23 & 79.42 & 78.31 & 78.96 & 77.62 & 78.58 & 77.79 & 76.89 & 75.24 \\
\cdashline{1-15}

\multirow{2}{*}{ChatLaw} 
 & 33B           & 81.82 & 81.23 & 80.09 & 80.89 & 80.64 & 79.84 & 78.73 & 79.32 & 78.01 & 78.99 & 77.24 & 78.96 & 75.68 \\
 & 13B           & 80.13 & 79.54 & 78.41 & 79.19 & 78.92 & 78.09 & 77.02 & 77.64 & 76.32 & 77.29 & 75.49 & 77.08 & 75.93 \\

\bottomrule 
\end{tabular}%
}

\label{tab:app1}
\end{table*}

\begin{table*}[htbp]
\centering
\caption{\small The performance (xCOMET-XL, \%) using \textbf{ChatGPT} as the Translation Command Module agent. Best results in each column are in \textbf{bold}. Model abbreviations: 13B-C: Baichuan-13B-Chat, 13B-B: Baichuan-13B-Base, 3-6B: ChatGLM3-6B, 2-6B: ChatGLM2-6B.}
\renewcommand{\arraystretch}{1.1} 
\setlength{\tabcolsep}{1.8pt}

\setlength{\aboverulesep}{1pt} 
\setlength{\belowrulesep}{1pt}

\setlength{\dashlinedash}{2pt} 
\setlength{\dashlinegap}{2pt}  

\resizebox{\textwidth}{!}{%
\tiny 
\begin{tabular}{ll ccc cc cc cc cc cc}
\toprule 
 & & \multicolumn{13}{c}{\textbf{Translation Command Module (ChatGPT)}} \\
\cmidrule(lr){3-15} 

\multicolumn{2}{l}{\textbf{Expert Review Module} $\downarrow$} & \multicolumn{13}{c}{\textbf{Translation Execution Module $\rightarrow$}} \\
\cmidrule(r){1-2} \cmidrule(lr){3-15}

 & & \multicolumn{3}{c}{OpenAI} 
 & \multicolumn{2}{c}{DeepSeek} 
 & \multicolumn{2}{c}{Qwen} 
 & \multicolumn{2}{c}{Baichuan} 
 & \multicolumn{2}{c}{ChatGLM} 
 & \multicolumn{2}{c}{ChatLaw} \\
\cmidrule(lr){3-5} \cmidrule(lr){6-7} \cmidrule(lr){8-9} \cmidrule(lr){10-11} \cmidrule(lr){12-13} \cmidrule(lr){14-15}

\textbf{Series} & \textbf{Model} & 
GPT-4o & GPT-4 & ChatGPT & 
MoE-16B & R1-Distill-32B & 
14B & 7B & 
13B-C & 13B-B & 
3-6B & 2-6B & 
33B & 13B \\
\midrule 

\multirow{3}{*}{OpenAI} 
 & \cellcolor{myblue}GPT-4o        & \cellcolor{myblue}\textbf{83.67} & \cellcolor{myblue}\textbf{83.34} & \cellcolor{myblue}\textbf{82.12} & \cellcolor{myblue}\textbf{82.89} & \cellcolor{myblue}\textbf{82.63} & \cellcolor{myblue}\textbf{81.82} & \cellcolor{myblue}\textbf{80.53} & \cellcolor{myblue}\textbf{81.06} & \cellcolor{myblue}\textbf{79.71} & \cellcolor{myblue}\textbf{80.24} & \cellcolor{myblue}\textbf{78.61} & \cellcolor{myblue}\textbf{79.31} & \cellcolor{myblue}\textbf{77.68} \\
 & GPT-4         & 83.42 & 83.01 & 81.88 & 82.57 & 82.36 & 81.54 & 80.22 & 80.79 & 79.38 & 79.92 & 78.23 & 78.94 & 77.29 \\
 & ChatGPT       & 82.69 & 82.13 & 81.04 & 81.68 & 81.42 & 80.61 & 79.56 & 79.91 & 78.64 & 79.23 & 77.58 & 78.12 & 76.66 \\
\cdashline{1-15}

\multirow{2}{*}{DeepSeek} 
 & MoE-16B            & 83.08 & 82.51 & 81.43 & 82.26 & 82.04 & 81.12 & 79.87 & 80.34 & 78.96 & 79.62 & 77.93 & 78.58 & 76.89 \\
 & R1-Distill-32B            & 82.84 & 82.22 & 81.18 & 82.03 & 81.87 & 80.94 & 79.71 & 80.19 & 78.82 & 79.46 & 77.78 & 78.43 & 76.74 \\
\cdashline{1-15}

\multirow{2}{*}{Qwen} 
 & 14B-Chat      & 82.33 & 81.76 & 80.68 & 81.52 & 81.24 & 80.57 & 79.36 & 79.78 & 78.41 & 79.08 & 77.34 & 77.92 & 76.27 \\
 & 7B-Chat       & 81.64 & 81.07 & 79.99 & 80.81 & 80.53 & 79.79 & 79.28 & 79.12 & 77.83 & 78.47 & 76.69 & 77.31 & 75.72 \\
\cdashline{1-15}

\multirow{2}{*}{Baichuan} 
 & 13B-Chat      & 81.96 & 81.32 & 80.27 & 81.09 & 80.86 & 80.04 & 78.98 & 80.08 & 78.13 & 78.82 & 77.01 & 77.63 & 75.98 \\
 & 13B-Base      & 81.03 & 80.49 & 79.42 & 80.17 & 79.89 & 79.16 & 78.08 & 78.67 & 78.34 & 77.89 & 76.14 & 76.71 & 75.06 \\
\cdashline{1-15}

\multirow{2}{*}{ChatGLM} 
 & 3-6B          & 81.62 & 80.97 & 79.91 & 80.68 & 80.39 & 79.62 & 78.54 & 79.16 & 77.91 & 78.98 & 76.62 & 77.23 & 75.54 \\
 & 2-6B          & 80.41 & 79.78 & 78.69 & 79.52 & 79.21 & 78.44 & 77.36 & 78.01 & 76.69 & 77.64 & 76.87 & 76.03 & 74.39 \\
\cdashline{1-15}

\multirow{2}{*}{ChatLaw} 
 & 33B           & 80.77 & 80.19 & 79.08 & 79.87 & 79.62 & 78.86 & 77.78 & 78.39 & 77.07 & 78.04 & 76.31 & 77.98 & 74.83 \\
 & 13B           & 79.09 & 78.53 & 77.42 & 78.18 & 77.93 & 77.12 & 76.08 & 76.71 & 75.39 & 76.36 & 74.58 & 76.13 & 75.02 \\

\bottomrule 
\end{tabular}%
}

\label{tab:app2}
\end{table*}

\clearpage

\onecolumn
\clearpage

\section{TransLaw Prompts}
\label{app:p}

\begin{figure}[H]
\vspace{-0.5cm}
    \centering
    \begin{tcolorbox}[promptbox, title={Prompt Template for Translation Command Agent ($\mathcal{A}_{\text{Com}}$)}]
    \small
    \textbf{Role:} You are a Senior Translation Project Manager with extensive experience in managing large-scale legal judgment translation (English-to-Traditional Chinese) workflows for the Hong Kong Department of Justice. Your responsibility is to orchestrate the entire workflow from input judgment to final translation output.

    \vspace{0.5em}
    \textbf{Task 1: Segmentation}
    Segment the raw English judgment into a sequence of independent sentences.
    \begin{itemize}[leftmargin=1.5em, label=-, nosep]
        \item \textbf{Note 1:} Do NOT split sentences within legal citations (e.g., [2025] HKCFA 8) or legislative references (e.g., Section 9 of the Theft Ordinance (Cap. 210)).
        \item \textbf{Note 2:} Preserve the logical flow. Do not break bullet points if they constitute a single semantic unit.
    \end{itemize}

    \vspace{0.5em}
    \textbf{Task 2: Context Management}
    Maintain a record of previously finalized translations to ensure terminological and logical consistency in the current sentence translation.

    \vspace{0.5em}
    \textbf{Task 3: Iterative Refinement}
    Manage the refinement loop based on expert feedback:
    \begin{itemize}[leftmargin=1.5em, label=-, nosep]
        \item \textbf{Integration:} Upon receiving feedback from the Expert Review Module, integrate the suggestions to update the current translation draft.
        \item \textbf{Termination:} If the feedback is empty or the maximum iteration limit is reached, mark the sentence as Finalized.
    \end{itemize}

    \vspace{0.5em}
    \textbf{Input Data (JSON):}
    \texttt{\{} \\
    \texttt{\ \ "task\_mode": "\{\{segmentation|refinement\}\}",} \\
    \texttt{\ \ "source\_text": "\{\{raw\_text\_or\_current\_draft\}\}",} \\
    \texttt{\ \ "feedback\_log": \{\{feedback\_json\_list\}\}} \\
    \texttt{\}}

    \vspace{0.5em}
    \textbf{Output Format (JSON):}
    \texttt{\{} \\
    \texttt{\ \ "status": "success",} \\
    \texttt{\ \ "segmented\_list": ["Sentence 1", "Sentence 2", ...],} // Populated if mode is segmentation \\
    \texttt{\ \ "updated\_draft": "Refined translation string..."} // Populated if mode is refinement \\
    \texttt{\}}
    \end{tcolorbox}
    \caption{Prompt template for Senior Translation Project Manager. It covers segmentation, context memory, iterative feedback integration, and final result aggregation.}
    \label{fig:prompt_cmd}
\end{figure}

\begin{figure}[H]
    \centering
    \begin{tcolorbox}[promptbox, title={Prompt Template for Legal Terminology Agent ($\mathcal{A}_{\text{Term}}$)}]
    \small
    \textbf{Role:} You are a HK Legal Terminologist working for the Department of Justice. You specialize in the precise retrieval and disambiguation of Common Law terms.

    \vspace{0.5em}
    \textbf{Task 1: Term Identification}
    Identify all Hong Kong legal terms (including Latin maxims, procedural terms, and Common Law terms of art) in the source sentence.

    \vspace{0.5em}
    \textbf{Task 2: RAG Consultation}
    Consult the "Retrieved Glossary Context" (RAG) derived from the Combined Department of Justice Glossaries of Legal Terms (https://www.glossary.doj.gov.hk/).

    \vspace{0.5em}
    \textbf{Task 3: Official Selection}
    Select the strictly official Traditional Chinese translation used in Hong Kong courts.

    \vspace{0.5em}
    \textbf{Task 4: Polysemy Disambiguation}
    Disambiguate polysemous terms (e.g., distinguishing "consideration" in contract law vs. general usage) based on the context.

    \vspace{0.5em}
    \textbf{Input Data:}
    \begin{itemize}[leftmargin=1.5em, label=-, nosep]
        \item Source Sentence ($s_i$): \texttt{\{\{s\_i\}\}}
        \item RAG Context: \texttt{\{\{Retrieved\_Glossary\_Entries\}\}}
    \end{itemize}

    \vspace{0.5em}
    \textbf{Output Format (JSON):}
    \texttt{\{} \\
    \texttt{\ \ "identified\_terms": [} \\
    \texttt{\ \ \ \ \{} \\
    \texttt{\ \ \ \ \ \ "src": "breach of statutory duty",} \\
    \texttt{\ \ \ \ \ \ "tgt": "違反法定責任",} \\
    \texttt{\ \ \ \ \}} \\
    \texttt{\ \ ]} \\
    \texttt{\}}
    \end{tcolorbox}
    \caption{Prompt template for HK Legal Terminologist, it ensures initial terms are retrieved from official glossary.}
    \label{fig:prompt_term}
\end{figure}

\begin{figure}[H]
    \centering
    \begin{tcolorbox}[promptbox, title={Prompt Template for Sentence Translation Agent ($\mathcal{A}_{\text{Trans}}$)}]
    \small
    \textbf{Role:} You are a Senior Court Translator with over 30 years of experience in the Hong Kong Judiciary. You are an expert in translating High Court judgments from English to Traditional Chinese.

    \vspace{0.5em}
    \textbf{Task 1: Term Integration and Translation}
    Generate a sentence-level translation by seamlessly injecting the mandatory terms into the target syntactic structure.
    \begin{itemize}[leftmargin=1.5em, label=-, nosep]
        \item \textbf{Strict Constraint:} You MUST use the exact translations provided in the \texttt{Term\_List}. Do not paraphrase or modify these verified legal expressions.
        \item \textbf{Syntactic Fluidity:} Ensure the resulting sentence structure is natural and grammatically correct in Traditional Chinese while accommodating the fixed terms.
    \end{itemize}

    \vspace{0.5em}
    \textbf{Task 2: Contextual Coherence}
    Ensure judicial coherence by referencing the \texttt{Global\_Context} (derived from preceding case facts and procedural history). The translation must maintain logical flow and consistent tone with previous sentences.

    \vspace{0.5em}
    \textbf{Task 3: Iterative Refinement (If Feedback Exists)}
    If \texttt{Review\_Feedback} is provided, modify the current draft to specifically address the identified errors (e.g., semantic misalignment, citation format) while preserving the correct parts.

    \vspace{0.5em}
    \textbf{Input Data (JSON):}
    \texttt{\{} \\
    \texttt{\ \ "source\_sentence": "\{\{text\}\}",} \\
    \texttt{\ \ "term\_list": \{\{json\_list\_from\_term\_agent\}\},} \\
    \texttt{\ \ "global\_context": "\{\{summary\_of\_preceding\_text\}\}",} \\
    \texttt{\ \ "review\_feedback": "\{\{optional\_feedback\_from\_reviewers\}\}"} \\
    \texttt{\}}

    \vspace{0.5em}
    \textbf{Output Format (JSON):}
    \texttt{\{} \\
    \texttt{\ \ "translation\_draft": "The draft translated sentence to be submitted to the Expert Review Module."} \\
    \texttt{\}}
    \end{tcolorbox}
    \caption{Prompt template for Senior Court Translator, it integrates mandatory terminology into the target syntactic structure to generate the translation candidate.}
    \label{fig:prompt_trans}
\end{figure}

\begin{figure}[H]
    \centering
    \begin{tcolorbox}[promptbox, title={Prompt Template for Semantic Alignment Agent ($\mathcal{A}_{\text{Align}}$)}]
    \small
    \textbf{Role:} You are a Senior Legal Reviser responsible for comparing the Source Sentence and the Translation Candidate to ensure accuracy.

    \vspace{0.5em}
    \textbf{Task: Error Detection}
    Check for the following errors:
    \begin{itemize}[leftmargin=1.5em, label=-, nosep]
        \item \textbf{Omission:} Is any key legal fact, date, or condition missing?
        \item \textbf{Hallucination:} Is there any added information not present in the source?
        \item \textbf{Logic:} Are subject-object relationships reversed (e.g., Plaintiff vs. Defendant)?
    \end{itemize}

    \vspace{0.5em}
    \textbf{Input Data (JSON):}
    \texttt{\{} \\
    \texttt{\ \ "source": "\{\{s\_i\}\}",} \\
    \texttt{\ \ "candidate": "\{\{draft\}\}"} \\
    \texttt{\}}

    \vspace{0.5em}
    \textbf{Output Format (JSON):}
    \texttt{\{} \\
    \texttt{\ \ "status": "PASS", // or "FAIL" \\}
    \texttt{\ \ "feedback": "Subject 'Appellant' was wrongly translated as 'Respondent'."} // null if PASS \\
    \texttt{\}}
    \end{tcolorbox}
    \caption{Prompt template for Senior Legal Reviser, it performs a bilingual check to detect semantic errors or omissions.}
    \label{fig:prompt_align}
\end{figure}

\begin{figure}[H]
    \centering
    \begin{tcolorbox}[promptbox, title={Prompt Template for Legal Term Review Agent ($\mathcal{A}_{\text{TermR}}$)}]
    \small
    \textbf{Role:} You are the Chief Terminologist responsible for validating that all terms in the candidate strictly match the official "Combined DOJ Glossaries".

    \vspace{0.5em}
    \textbf{Task: Glossary Verification}
    \begin{itemize}[leftmargin=1.5em, label=-, nosep]
        \item \textbf{Strict Compliance:} Ensure no layman terms are used (e.g., use "合約" not "合同").
        \item \textbf{Consistency:} Ensure the same term is not translated differently within the same context.
    \end{itemize}

    \vspace{0.5em}
    \textbf{Input Data (JSON):}
    \texttt{\{} \\
    \texttt{\ \ "candidate": "\{\{draft\}\}",} \\
    \texttt{\ \ "glossary\_db": "\{\{db\_access\}\}"} \\
    \texttt{\}}

    \vspace{0.5em}
    \textbf{Output Format (JSON):}
    \texttt{\{} \\
    \texttt{\ \ "status": "FAIL",} \\
    \texttt{\ \ "feedback": "Term 'consideration' must be translated as '代價', not '考慮'."} \\
    \texttt{\}}
    \end{tcolorbox}
    \caption{Prompt template for Chief Terminologist, it validates lexical accuracy against the authoritative government glossary.}
    \label{fig:prompt_termr}
\end{figure}

\begin{figure}[H]
    \centering
    \begin{tcolorbox}[promptbox, title={Prompt Template for Legal Citation Agent ($\mathcal{A}_{\text{Cita}}$)}]
    \small
    \textbf{Role:} You are a Professional Legal Editor specializing in HKCFA judgment citation standards.

    \vspace{0.5em}
    \textbf{Task: Format Validation}
    \begin{itemize}[leftmargin=1.5em, label=-, nosep]
        \item \textbf{Case References:} Check italicization and brackets. (e.g., HKSAR v. Chan [2025] HKCFA 8).
        \item \textbf{Legislation:} Ensure correct use of "Cap." and "Section". (e.g., Theft Ordinance (Cap. 210)).
    \end{itemize}

    \vspace{0.5em}
    \textbf{Input Data (JSON):}
    \texttt{\{} \\
    \texttt{\ \ "candidate": "\{\{draft\}\}"} \\
    \texttt{\}}

    \vspace{0.5em}
    \textbf{Output Format (JSON):}
    \texttt{\{} \\
    \texttt{\ \ "status": "PASS",} \\
    \texttt{\ \ "feedback": null} \\
    \texttt{\}}
    \end{tcolorbox}
    \caption{Prompt template for Professional Legal Editor, it enforces rigid citation formatting standards for cases and legislation.}
    \label{fig:prompt_cita}
\end{figure}

\begin{figure}[H]
    \centering
    \begin{tcolorbox}[promptbox, title={Prompt for Stylistic Fidelity Polishing Agent ($\mathcal{A}_{\text{StyleP}}$)}]
    \small
    \textbf{Role:} You are a Senior Judicial Editor responsible for refining the text to uphold the authority of the Court.

    \vspace{0.5em}
    \textbf{Task: Tone and Fluency}
    \begin{itemize}[leftmargin=1.5em, label=-, nosep]
        \item \textbf{Judicial Tone:} The text must sound authoritative and dignified, exactly like a High Court judgment (e.g., use "陳詞" for "submit", "裁定" for "hold").
        \item \textbf{Fluency:} Fix choppy sentences to ensure the legal arguments connect logically and read smoothly in Traditional Chinese.
    \end{itemize}

    \vspace{0.5em}
    \textbf{Input Data (JSON):}
    \texttt{\{} \\
    \texttt{\ \ "candidate": "\{\{draft\}\}"} \\
    \texttt{\}}

    \vspace{0.5em}
    \textbf{Output Format (JSON):}
    \texttt{\{} \\
    \texttt{\ \ "final\_text": "The official Traditional Chinese judgment text awaiting publication"} \\
    \texttt{\}}
    \end{tcolorbox}
    \caption{Prompt template for Senior Judicial Editor, it refines the linguistic output to ensure judicial authority and fluency.}
    \label{fig:prompt_style}
\end{figure}

\end{CJK*}

\clearpage

\clearpage

\section{Evaluation Guidelines for Human Experts}
\label{app:guide}
\begin{CJK*}{UTF8}{bsmi}

\begin{figure}[H]
    \centering
    \begin{tcolorbox}[promptbox, breakable, title={Evaluation Guidelines for Human Experts}]
    \normalsize
    
    \textbf{General Instructions:} Evaluators must give each translation a score between 0 and 10. Evaluators are provided with the source text, the ``gold translation'' (official court translation), and the predicted translation.
    
    \vspace{0.5em}
    \textbf{Score Criteria:} Scores shall reflect the completeness and accuracy of the predicted translation.
    
    \vspace{0.5em}
    \textbf{Point Deduction System:}
    \begin{itemize}[leftmargin=1.5em, nosep]
        \item \textbf{Score 10:} A perfectly complete and accurate translation.
        \item \textbf{-1 Point:} Relevant legal term translated in an unusual but correct manner; minor stylistic lapse.
        \item \textbf{-2 Points:} Legally relevant term translated erroneously; Relevant term missing.
        \item \textbf{-4 Points:} Critical errors (e.g., referencing the wrong statute or offense).
    \end{itemize}
    
    \vspace{0.5em}
    \textbf{Examples:}
    \begin{enumerate}[leftmargin=1.5em, label=\arabic*), nosep]
        \item \textbf{Score 10}: Perfect match in legal terminology and meaning.\\
        \textit{Source:} Mr. Hu referred to a number of decided cases involving the offence of wounding with intent to support his contention that, in the circumstances of the present case, 3 years’ imprisonment was manifestly excessive.\\
        \textit{Gold:} 胡大律師引用多宗有關「有意圖而傷人」的案件，指以本案案情而言，3 年監禁刑期是明顯過重。\\
        \textit{Prediction:} 胡大律師引用多宗有關「有意圖而傷人」的案件，指以本案案情而言，3 年監禁刑期是明顯過重。
        
        \vspace{0.5em}
        \item \textbf{Score 9}: (-1 for unusual translation of ``manifestly excessive'')\\
        \textit{Prediction:} 胡大律師引用多宗有關「有意圖而傷人」的案件，指以本案案情而言，3 年監禁刑期\textbf{實在太多了}。\\
        \textit{Note:} ``Too much'' (實在太多了) is colloquial; standard term is ``manifestly excessive'' (明顯過重).
        
        \vspace{0.5em}
        \item \textbf{Score 6}: (-4 for critical error in legal offense)\\
        \textit{Prediction:} 胡大律師引用多宗有關\textbf{誤殺}的案件，指以本案案情而言，3 年監禁刑期是明顯過重。\\
        \textit{Note:} Critical error: ``Wounding with intent'' mistranslated as ``Manslaughter'' (誤殺).
    \end{enumerate}
    
    \end{tcolorbox}
    \caption{The detailed evaluation guidelines provided to human experts.}
    \label{fig:annotation_guidelines}
\end{figure}

\end{CJK*}

\end{document}